\documentclass[lettersize,journal,onecolumn]{IEEEtran}

\ifCLASSINFOpdf
\usepackage[pdftex]{graphicx}
  \graphicspath{{../pdf/}{../jpeg/}}
  \DeclareGraphicsExtensions{.pdf,.jpeg,.png}
\else
  \usepackage[dvips]{graphicx}
  \graphicspath{{../eps/}}
  \DeclareGraphicsExtensions{.eps}
\fi
\usepackage[switch]{lineno}
\usepackage{ulem}
\usepackage{cancel}

\usepackage{epstopdf}
\usepackage{amsmath,amsthm, amssymb}
\usepackage{mathrsfs}
\usepackage{threeparttable}
\usepackage{bm}
\usepackage{multirow}
\usepackage{booktabs}
\usepackage{color}
\usepackage{tabularx}
\usepackage{setspace}
\usepackage{verbatim}
\usepackage{stfloats}
\usepackage{caption}
\usepackage{float}
\usepackage[T1]{fontenc}
\usepackage{mathrsfs}
\usepackage{soul}
\usepackage{cite}
\usepackage{color}
\usepackage{multirow}

\usepackage{stfloats}
\usepackage{mathtools}
\usepackage{amsmath}
\allowdisplaybreaks[0]
\usepackage{subfigure}

\usepackage{fancyhdr}
\usepackage{cancel}
\usepackage{array}

\usepackage{xcolor}
\makeatletter
\newif\if@restonecol
\makeatother

\usepackage[linesnumbered,ruled,vlined]{algorithm2e}
\usepackage{algpseudocode}
\usepackage{amsmath}
\usepackage{color}

\let\oldequation\equation
\let\oldendequation\endequation
\renewenvironment{equation}
  {\linenomathNonumbers\oldequation}
  {\oldendequation\endlinenomath}

\begin{document}
\normalem
\title{Cooperative Edge Caching Based on Elastic Federated and Multi-Agent Deep Reinforcement Learning in Next-Generation Networks }

\author{Qiong Wu,~\IEEEmembership{Senior Member,~IEEE}, Wenhua Wang, Pingyi Fan, ~\IEEEmembership{Senior Member,~IEEE}, Qiang Fan, \\Huiling Zhu, ~\IEEEmembership{Senior Member,~IEEE} and Khaled B. Letaief,~\IEEEmembership{Fellow,~IEEE}

\thanks{
Qiong Wu and Wenhua Wang are with the School of Internet of Things Engineering, Jiangnan University, Wuxi 214122, China, and also with the State Key Laboratory of Integrated Services Networks (Xidian University),  Xi'an 710071, China (e-mail: qiongwu@jiangnan.edu.cn, wenhuawang@stu.jiangnan.edu.cn)

Pingyi Fan is with the Department of Electronic Engineering, Beijing National Research Center for Information Science and Technology, Tsinghua University, beijing 100084, China (Email: fpy@tsinghua.edu.cn)

Qiong Fan is with Qualcomm, San Jose, CA 95110, USA (e-mail: qf9898@gmail.com)

Huiling Zhu is with the School of Engineering, University of Kent, CT2 7NT Canterbury, U.K.(e-mail:H.Zhu@kent.ac.uk)

K. B. Letaief is with the School of Engineering, Hong Kong University of Science and Technology (HKUST), Hong
Kong (email:eekhaled@ust.hk)
}

}


{}
\maketitle
\begin{abstract}
Edge caching is a promising solution for next-generation networks by empowering caching units in small-cell base stations (SBSs), which allows user equipments (UEs) to fetch users' requested contents that have been pre-cached in SBSs. It is crucial for SBSs to predict accurate popular contents through learning while protecting users' personal information. Traditional federated learning (FL) can protect users' privacy but the data discrepancies among UEs can lead to a degradation in model quality. Therefore, it is necessary to train personalized local models for each UE to predict popular contents accurately. In addition, the cached contents can be shared among adjacent SBSs in next-generation networks, thus caching predicted popular contents in different SBSs may affect the cost to fetch contents. Hence, it is critical to determine where the popular contents  are cached cooperatively. To address these issues, we propose a cooperative edge caching scheme based on elastic federated and multi-agent deep reinforcement learning (CEFMR) to optimize the cost in the network. We first propose an elastic FL algorithm to train the personalized model for each UE, where adversarial autoencoder (AAE) model is adopted for training to improve the prediction accuracy, then a popular content prediction algorithm is proposed to predict the popular contents for each SBS based on the trained AAE model. Finally, we propose a multi-agent deep reinforcement learning (MADRL) based algorithm to decide where the predicted popular contents are collaboratively cached among SBSs. Our experimental results demonstrate the superiority of our proposed scheme to existing baseline caching schemes.



\end{abstract}

\IEEEpeerreviewmaketitle
\section{Introduction}
\label{sec1}
\IEEEPARstart{I}{n} recent years, with the increasing popularity of smart devices, we have witnessed an unprecedented growth in mobile data traffic, which has imposed a heavy burden on wireless networks\cite{1800286,3161028,Li2578925}. As users increasingly rely on user equipment (UEs) like mobile devices and home routers to access content from wireless networks, it becomes challenging for ensuring a satisfactory quality of service to meet their demands\cite{Zhang060238, Wu80211, Yao1207121}.
To address this challenge, edge caching has emerged as a promising solution for next-generation networks \cite{6871674,3198074,zhang56905703}.
Through the implementation of caching units in wireless edge nodes, such as small-cell base stations (SBSs), UEs can fetch users' desired contents from nearby SBSs instead of remote servers or cloud. This process significantly reduces traffic load, alleviates network congestion, reduces latency and improves system performance\cite{2577278}.

To  efficiently enable UEs to fetch users' desired contents from SBSs, SBSs need to predict popular contents based on the interests of users within their coverage area and proactively cache  these popular contents in advance.
Due to the unique preference of each user, the popular contents cached in different SBSs may exhibit surprising variations \cite{7537172}. Machine learning (ML) can overcome this issue by training  based on users' data to extract hidden features, thereby effectively predicting popular contents \cite{Yan2019}.
However, most traditional ML algorithms need to train and analyze the data generated by users, which may involve personal sensitive information.
Federated Learning (FL) has emerged as a potential solution that can protect the privacy and security of users' data \cite{9378161,3172370}.
For the traditional FL, each UE trains its local model based on its users' data, and then only uploads the trained local model to the local SBS instead of its training data. Then the local SBS aggregates the uploaded models from UEs to update the global model.
However, traditional FL tends to generalize user preferences, neglecting the personalized patterns of individual UEs. This general approach may not cater to the specific needs and behaviors of individual UEs, leading to potential inefficiencies in someway. To address this limitation, we introduce the elastic FL algorithm which adjusts the learning process dynamically based on the unique characteristics of each UE. By assigning specific weights to each local model based on differences between the global and local models, one can capture users' individual characteristics\cite{3053055}. This elastic adjustment ensures that the final model is both globally informed and locally sensitive. This dual feature significantly enhances both prediction accuracy and caching hit ratios in edge caching scenarios.

After each local SBS has predicted the popular contents, deploying edge caches poses an additional challenge.  
In next-generation networks, the importance of collaborative caching is further strengthened. Here, cached contents can be shared among adjacent SBSs via the Xn interface \cite{NR2019}. This allows UEs to indirectly obtain the requested contents from the adjacent SBSs even if they cannot directly fetch the requested contents from the local SBS.
The majority of research in collaborative caching has focused on joint optimization with metrics like resource allocation and latency reduction, rather than concentrating on popular contents. Additionally, while studies employing artificial intelligence technology generally use popular contents in their caching strategies, they often overlook the crucial aspect of predicting content popularity for future proactive caching. Therefore, predicting future popular contents and utilizing them for collaborative caching can significantly enhance caching efficiency.
However, the number of predicted popular contents for each SBS is usually larger than the limited cache capacity of the SBS. Therefore, this requires cooperation between SBSs to cache popular contents. Additionally, the placement of these popular contents in the right SBSs is also critical, as it directly impacts the cost of fetching contents across the network. In addition, the dimensions of caching decision will increase with the number of SBSs and cached contents, inevitably adding complexity to the caching problem in the entire network. This makes it challenging to coordinate caching among SBSs to reduce the cost for fetching contents in the next-generation network.
Multi-agent reinforcement learning (MADRL) is a suitable method that can overcome the above problem, and has begun to be used in cooperative edge caching. Instead of the traditional way of making separate decisions, MADRL promotes cooperation among SBSs. Each SBS, which is considered an agent, does not just decide where to cache popular contents based on its own prediction. Instead, it should work together with other SBSs, taking into account their decisions to create a network-wide caching strategy driven by content popularity. If some UEs' demands change, MADRL can adjust the caching decisions. This adaptability is crucial for handling the increased complexity that comes with the growing caching needs in the next-generation network.


In this paper, we propose a cooperative edge caching scheme based on elastic federated learning and multi-agent deep reinforcement learning (CEFMR) scheme to optimize the cost of fetching contents in the next-generation network.
The contributions of this paper are summarized as follows\footnote{This paper has been submitted to IEEE TNSM. The source code has been released at: https://github.com/qiongwu86/Edge-Caching-Based-on-Multi-Agent-Deep-Reinforcement-Learning-and-Federated-Learning}

\begin{itemize}
\item[1)] Each SBS employs an adversarial autoencoder (AAE) model to predict the content popularity within its own coverage. It helps to learn deep latent representations from users' historical requests and contextual information, enabling the discovery of implicit relationships between users and  contents, thus improving prediction accuracy.

\item[2)] We propose an elastic FL algorithm to train the AAE  model for each SBS, which assigns specific weight for each UE's model based on the differences between the global model and local model, thereby it can protect users' privacy and obtain the personalized local model for each UE.

\item[3)] We formulate the collaborative edge caching problem as a caching decision problem based on the MADRL framework, where states, actions, and reward have been defined. This formulation aims to optimize the cost for fetching contents in the network. Afterwards, we propose the multi-agent deep deterministic policy gradient (MADDPG) based algorithm to learn the optimal caching decision.

\end{itemize}

The remainder of this paper is structured as follows.
Section \ref{sec2} reviews the related works on the research on popularity prediction and collaborative caching.
Section \ref{System model} provides a concise overview of the system model.
Section \ref{Content Popularity Predicition Algorithm} introduces the proposed cooperative edge caching scheme.
We present some simulation results in Section \ref{sec6}, and then conclude them in Section \ref{sec7}.

\section{Related Work}
\label{sec2}
We first reviewed the research on popularity prediction and collaborative caching, and then we reviewed the studies on collaborative caching based on MADRL and FL.

Currently, there are studies on content popularity prediction.
In \cite{Pervej9322208}, Pervej \textit{et al.} proposed a novel approach to edge caching in next-generation wireless networks using a long short-term memory based sequential model to dynamically predict user preferences for content caching. Additionally, they formulated a non-convex optimization problem to minimize content sharing costs and employed a greedy algorithm to find a sub-optimal solution.
In \cite{Yao201900115}, Yao \textit{et al.} introduced vehicular content centric network to address mobility challenges in vehicular ad hoc networks by implementing content centric networking. They proposed popularity-based content caching, a cache replacement scheme that predicts content popularity using hidden Markov Model based on received interests, request ratio, frequency and content priority. 
In \cite{Tao9685947}, Tao \textit{et al.} proposed a Gaussian process based Poisson regressor to accurately and efficiently model content request patterns. Bayesian learning is employed to robustly learn model parameters, and to handle the lack of closed-form expression for the posterior distribution, a Stochastic Variance Reduced Gradient Hamiltonian Monte Carlo method is applied. 
In \cite{Liu2901525}, Liu \textit{et al.} presented a content popularity prediction algorithm based on auto-regressive models for Information-Centric Networking in the Internet of Things, improving cache hit rates, reducing network traffic, and minimizing service access delays, particularly beneficial for real-time streaming media services.
In \cite{Feng10152582}, Feng \textit{et al.} introduced a content popularity prediction algorithm, GRU-Attention, tailored for Vehicle Named Data Networking, which relied on in-network caching to swiftly serve subsequent content requests. By leveraging the attention mechanism and GRU model, it accurately forecasted future popular contents based on multiple historical request patterns, enhancing prediction accuracy and optimizing cache utilization within the network.

There are already studies on collaborative caching.
In \cite{Pervej9322101}, Pervej \textit{et al.} tackled the optimization of collaborative caching in heterogeneous edge networks by developing an improved Particle Swarm Optimization algorithm, demonstrating through numerical analysis and simulations that this approach significantly enhances cache hit ratios compared to traditional caching strategies.
In \cite{Saputra9013745}, Saputra \textit{et al.} employed a cooperative mobile edge caching network by jointly optimizing collaborative caching and routing decisions to reduce content access delays and backhaul traffic. To address the optimization challenge, they transformed the problem into a mixed-integer nonlinear programming problem and utilized a branch-and-bound algorithm with the interior-point method to find a near-optimal solution.
In \cite{Xie8891441}, Xie \textit{et al.} proposed a cooperative caching scheme where user terminals cache popular services to an intelligent routing relay by using an online reverse auction combined with a first-come-first-served strategy.
In \cite{Li2875060}, Li \textit{et al.} employed a crowdsourced approach for cooperative caching in Content-Centric Mobile Networking. They introduced a robust caching control scheme using Kullback-Leibler divergence to manage uncertainties in user mobility, finally formulated as a chance-constrained robust optimization problem.
In \cite{Liao8254854}, Liao \textit{et al.} employed cooperative caching strategies using maximum distance separable codes for content restructuring and optimizes both content placement and cooperation policies among base stations. A compound caching strategy, namely multicast-aware cooperative caching assuming fixed and dynamic cooperative policies, respectively, was developed to combine the merits of multicast-aware content delivery and cooperative content sharing.
In \cite{Fang3001229}, Fang \textit{et al.} investigated the economic interactions between Internet service providers and content providers in wireless networks, focusing on optimizing content delivery through edge caches and content popularity. They proposed a centralized model to solve the profit split problem and employed a Stackelberg game for a distributed solution.

Furthermore, there are already studies on collaborative caching based on MADRL.
In \cite{Araf2023}, Araf \textit{et al.} employed a multi-agent actor-critic reinforcement learning approach, combined with UAV (Unmanned Aerial Vehicle) assistance, to tackle the problem of optimizing cooperative caching strategies on the network edge. 
In \cite{Chen3044298}, Chen \textit{et al.} proposed a multi-agent actor-critic framework with variational recurrent neural networks (VRNN) to estimate content popularity. Through multi-agent deep reinforcement learning, the cooperative edge caching algorithm substantially improved the benefits of edge caches in ultra-dense next-generation networks.
In \cite{2894403}, Jiang \textit{et al.} employed multi-agent reinforcement learning to design content caching strategies in mobile device-to-device (D2D) networks, addressing the challenge of content caching without prior knowledge of popularity distribution. The proposed approach, using Q-learning, maximized caching rewards by treating users as agents and caching decisions as actions.
In \cite{3066458}, Wu \textit{et al.} proposed the integration of cooperative coded caching using maximum distance separable (MDS) codes in small cell networks to alleviate traffic loads. To adapt to dynamic and unknown content popularity, a multi-agent deep reinforcement learning framework was introduced to intelligently update cached content, with a focus on minimizing long-term expected front-haul traffic loads, resulting in improved caching efficiency.
In \cite{2968326}, Zhong \textit{et al.} investigated cache placement by using the collaborative MADRL algorithm in ultra-dense small cell networks (SCNs) to maximize cache hit rates in centralized and decentralized scenario settings.
In \cite{00062}, Jiang \textit{et al.} explored edge caching for the internet of vehicles (IoVs) using distributed MARL. They introduced a hierarchical caching architecture to minimize content delivery overhead, extended single-agent reinforcement learning to a multi-agent system, and achieved superior performance in comparison to other caching strategies through simulation results.
In \cite{2713384}, Song \textit{et al.} addressed the content caching and sharing problem in cooperative base stations while considering unknown content popularity distribution and content retrieval costs. This paper formulated the problem as a multi-armed bandit (MAB) learning problem, jointly optimizing caching and sharing strategies. Two algorithms, a centralized approach using semidefinite relaxation and a decentralized approach based on the alternating direction method of multipliers (ADMM), were proposed to solve the problem efficiently.
In \cite{2023.03.004}, Cai \textit{et al.} proposed a cross-tier cooperative caching architecture for all content that allows distributed cache nodes to cooperate, using a MADRL approach to model the decision process between heterogeneous cache nodes.
In \cite{9155373}, Wang \textit{et al.} introduced an intelligent edge caching framework named MacoCache, which incorporated MADRL to enhance edge caching performance in distributed environments. By employing the advanced actor-critic method and integrating long short-term memory (LSTM) for time series dynamics, MacoCache successfully reduced latency by an average of 21\% and costs by 26\% compared to existing learning-based caching solutions.
In \cite{2022.00.006}, Chen \textit{et al.} utilized a refined MADRL tailored for the 5G network structure to improve scalable video coding (SVC)-based caching and video quality service strategies. The introduced value-decomposed dimensional network (VDDN) algorithm and a dimension decomposition method in the dueling deep Q-network addressed the challenges of large action spaces and computational efficiency.
In \cite{3264553}, Zhang \textit{et al.} introduced an edge caching model for IoV content distribution, allowing vehicles to collaboratively cache content and account for varying content popularity and channel conditions. An autonomous decision-making approach, an edge caching approach for the IoV based on multi-agent deep reinforcement learning, was proposed where vehicles act as agents to optimize caching decisions.
However, most previous studies have primarily used MADRL to tackle issues in collaborative content caching. It is important to point out that these studies often did not address the more complex challenges associated with predicting what contents will be popular in the future. Specifically, they did not focus on the intricacies of using machine learning to accurately forecast content popularity, which is crucial for effective content management.

Recently, there have been several works on designing content caching schemes based on MADRL and FL.
In \cite{10039249}, Chang \textit{et al.} introduced a federated multi-agent reinforcement Learning (FMRL) method for optimizing edge caching. Using a Markov decision process (MDP) and a specialized regression model, it predicted content popularity, improved server rewards, and optimized cache decision-making. An enhanced FL technique aggregated local models, improving training and preserving user privacy.
In \cite{3133291}, Prathiba \textit{et al.} introduced the federated Learning and edge cache-assisted Cybertwin framework in response to rising autonomous vehicle (AV) technology and mobile traffic. This framework integrated the cybertwin concept to merge physical and digital systems. It employed the federated MADRL algorithm for optimized learning, considering various factors. Additionally, the federated reinforcement learning-based edge caching algorithm was used for efficient caching in 6G vehicle-to-everything networks. 
In \cite{3235443}, Sun \textit{et al.} employed FL to perceive user preferences across various districts without exposing user requests. A content blocks-based RL is utilized for intelligent caching, taking into account the receiving scenarios of AVs. By integrating FL with MARL, this paper achieved distributed training of local models, which not only reduced the state-action space dimension but also circumvented local optimal solutions.
However, the previous studies did not take into account the personalization needs of individual UE in FL training. Additionally, they overlooked the fact that different SBSs may have various popular contents, which can significantly impact caching strategies.

In conclusion, previous research has not adequately addressed two key aspects in designing content caching schemes. Firstly, FL approaches have not considered the personalized and elastic requirements of each UE, which is essential for optimizing user-specific performance. Secondly, MADRL methods have largely ignored the importance of predicting popular contents before making collaborative caching decisions. These gaps in the literature have motivated our research, as we aim to develop personalized models for UEs while also focusing on proactive collaborative caching strategies based on predicted content popularity.

\begin{figure*}
	\center
	\includegraphics[scale=1]{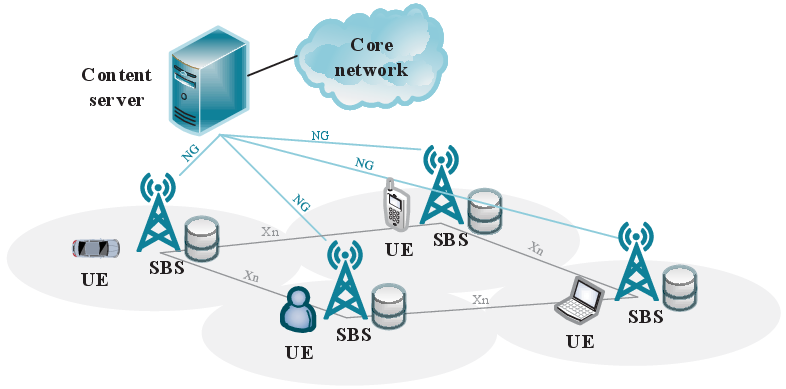}
	\caption{Cooperative edge caching system}
	\label{fig1}
	\vspace{-0.5cm}
\end{figure*}

\begin{table*}
	\footnotesize
	\caption{Main notations.}
	\label{tab1}
	\centering
	\begin{tabular}{|p{1cm}<{\centering}|p{5.8cm}|p{1cm}<{\centering}|p{5.8cm}|}	\hline
		\textbf{Notation} &\textbf{Description} &\textbf{Notation} &\textbf{Description}\\
		\hline
		$\boldsymbol{a^t}$ & the global action at time slot $t$. &$a_b^t$ & the action of SBS $b$ at time slot $t$.\\
		\hline
		$B$ & the number of SBSs. &$b_d$ & the bias vector of the decoder network.\\
		\hline
		$b_e$ & the bias vector of the encoder network. & $C$ & the cache capacity of each SBS.\\
		\hline
		$c_b^t$ &\multicolumn{1}{m{6cm}|}{the current contents cached by SBS $b$ at time slot $t$.} & $d$ &\multicolumn{1}{m{6cm}|}{the size of entire data of all UEs within the coverage of SBS.}\\
		\hline
		$d(\cdot)$ & \multicolumn{1}{m{6cm}|}{the non-linear and logically-activated function of the discriminator network.} &$d_i$ & \multicolumn{1}{m{6cm}|}{the size of the training data of each UE.}\\
		\hline
		$D$ & the replay buffer. &$D(z)$ & the confidence of $z$.\\
		\hline
		$e$ & \multicolumn{1}{m{6cm}|}{the number of iteration of the local AAE training for UE.} &$E^{\prime}$ & \multicolumn{1}{m{6cm}|}{the number of episodes in the testing phase of MADRL algorithm.}\\
		\hline
		$F_p$ & \multicolumn{1}{m{6cm}|}{the number of predicted popular contents.} &$J(\boldsymbol{\pi^t})$ & \multicolumn{1}{m{6cm}|}{the expected long-term discounted reward at time slot $t$.}\\
		\hline
		$|L|$ & the number of layers of the AAE model. &$M$ & the number of transition tuples in a mini-batch.\\
		\hline
		$n_{i,k}^r$ & \multicolumn{1}{m{6cm}|}{local training data set of UE $i$ in the iteration $k$ of round $k$.} & $p(\cdot)$ & \multicolumn{1}{m{6cm}|}{the non-linear and logically-activated function of the decoder network.} \\
		\hline
		$P(z)$ & the predefined distribution. &$p_b$ & \multicolumn{1}{m{6cm}|}{the predicted popular contents of SBS $b$ coverage.}\\
		\hline
		$q(\cdot)$ & \multicolumn{1}{m{6cm}|}{the non-linear and logically-activated function of the encoder network.} & $R^t$ & \multicolumn{1}{m{6cm}|}{the global reward  at the time slot $t$.}\\
		\hline
		$R_b^t$ & \multicolumn{1}{m{6cm}|}{the local reward of SBS $b$ at time slot $t$.} & $R_{max}$ & \multicolumn{1}{m{6cm}|}{the number of rounds to train the AAE model in the elastic FL.} \\
		\hline
		$r_b^r(t)$ & \multicolumn{1}{m{6cm}|}{the number of contents that SBS $b$ fetches from the local cache at time slot $t$.} &$r_{b, c}^r(t)$ & \multicolumn{1}{m{6cm}|}{the number of contents that SBS $b$ fetches from CS at time slot $t$.}\\
		\hline
		$r_{b, e}^r(t)$ & the contents replaced in SBS $b$ during time slot $t$. &$r_{b, n}^r(t)$ & \multicolumn{1}{m{6cm}|}{the number of contents that SBS $b$ fetches from adjacent SBSs at time slot $t$.}\\
		\hline
		$s_b^t$ & the local state of SBS $b$ at time slot $t$. &$\boldsymbol{s^t}$ & the global state at time slot $t$.\\
		\hline
		$T$ & the number of time slots of the MADRL algorithm. &$X_i^r$ & the rating matrix of UE $i$ in round $r$.\\
		\hline
		$\tilde{X}_i^r$ & the reconstructed rating matrix of UE $i$ in round $r$. &$\alpha$ & \multicolumn{1}{m{6cm}|}{the cost of fetching the contents from local SBS $b$.}\\
		\hline
		$\alpha_i$ & the elastic parameter of UE $i$. &$\beta$ & \multicolumn{1}{m{6cm}|}{the cost that the UE within the coverage of local SBS $b$ fetch the contents from adjacent SBSs.}\\
		\hline
		$\chi$ & \multicolumn{1}{m{6cm}|}{the cost of fetching the contents from CS.} &$\delta$ & the cost of replacing a content.\\
		\hline
		$\gamma$ & the discount factor. &$\eta$ & the fixed learning rate.\\
		\hline
		$\theta_b$ & \multicolumn{1}{m{6cm}|}{the parameters of the actor network of agent $b$.} &$\theta_b^{\prime}$ & \multicolumn{1}{m{6cm}|}{the parameters of the target critic network of agent $b$.}\\
		\hline
		$\varphi$ & the parameters of the global critic network. &$\varphi^{\prime}$ & the parameters of the target global critic network.\\
		\hline
		$\tau$ & the constant is used to update the target networks. &$\omega^r$ & the global AAE model in round $r$.\\
		\hline
		$\omega_e^r$ & the global encoder network in round $r$. &$\omega_d^r$ & the global decoder network in  round $r$.\\
		\hline
		$\omega_{d'}^r$ & the global discriminator network in  round $r$. &$\omega_i^r$ & \multicolumn{1}{m{6cm}|}{the local model of UE $i$ in  round $r$.}\\
		\hline
	\end{tabular}
\end{table*}

\section{System Model}
\label{sec3}


\label{System model}
As illustrated in Fig. \ref{fig1}, we consider a cooperative edge caching system in the next-generation network consisting of a content server (CS), a set of SBSs denoted as $\mathbb{B}=\{1, \ldots, b, \ldots, B\}$, where $B$ is the number of SBSs, and a certain number of UEs.
Each UE fetches the requested contents from SBSs or CS for multiple users, and stores a large amount of users' data, i.e., local data.
Each data is a representation in vector form that encompasses various aspects of users' information such as the users' identification number (ID), gender, age, the contents that users may request, and users' ratings for the contents that they have previously requested. The rating is between $0$ and $1$ which can measure a user's interest in previously requested content, and this rating method is common in many research and application scenarios \cite{3017474,307319}. $0$ represents that the user is not interested in this requested content or the user has not requested this content before, while $1$ represents that the user is most interested in this requested content. Each UE randomly selects a portion of local data to establish a training set, and the remaining data is utilized as a test set.
The CS is linked to the core network by a backhaul link, and it stores all contents that users can request. Each SBS is connected to the CS through a backhaul link connected by the next-generation (NG) interface. SBSs are deployed at the network edge close to the UEs. 
The time duration is divided into time slots. In each time slot, each UE collects contents that users want to request and generates a requested information. This requested information may be a small-sized list containing the contents labels which users desire. Therefore, the requested information will not occupy too much resources. Then UEs send these requested information to their local SBSs to fetch the requested contents.
The cache capacity of a single SBS is actually very limited, the adjacent SBSs can share cached contents with each other through a wired-link connected by the Xn interface. Hence, a UE can fetch a requested content from the local SBS, adjacent SBSs or CS. Specifically,





\subsubsection{Local SBS}If the requested content is cached in the local SBS, the UE can fetch the requested content directly from the local SBS.
\subsubsection{Adjacent SBS}If the requested content is not cached in the local SBS but an adjacent SBS caches the request content, the local SBS can fetch the requested content from the adjacent SBS and then deliver it to the UE.
	If multiple adjacent SBSs have all cached the requested content, the local SBS randomly selects one of the adjacent SBSs to fetch the requested content.

\subsubsection{CS}
If both the local SBS and adjacent SBSs do not cache the requested content, the local SBS fetches the requested content from the CS and then delivers it to the UE.



In order to enable UEs within the coverage of each SBS to fetch requested contents more effectively, each SBS needs to accurately predict the popular contents. These popular contents are predicted based on users' ratings for the contents. However, it is difficult to distinguish whether the user is interested in this content or has requested it before if a rating for one content is 0. Therefore, it is necessary to predict and reconstruct the ratings of each content. The AAE model can effectively extract hidden features of ratings and reconstruct the ratings for each content. Hence the reconstructed ratings will contain fewer $0$. This helps to predict the popular contents more accurately. Thus the AAE model is adopted as the local model and global model in the FL algorithm and the popular contents for each SBS are predicted based on the trained AAE model. The proposed CEFMR scheme is described as follows. We first propose an elastic FL algorithm which can assign specific weight for each local model based on the differences between the global model and local model to train the personalized local model for each UE. Then we propose a popular content prediction algorithm to predict the popular contents for each SBS based on the trained AAE model. Finally, due to the limited cache capacity of each SBS, we construct a MADRL framework and then adopt the MADDPG algorithm to decide where the predicted popular contents are collaboratively cached among SBSs to reduce the cost for fetching contents. We will elaborate the CEFMR scheme in the next section. The main notations are listed in Table I.

\section{Cooperative Edge Caching Scheme}
\label{Content Popularity Predicition Algorithm}

In this section, we will introduce the CEFMR cooperative edge caching scheme. We first propose an elastic FL algorithm to train the personalized AAE model for each UE, then a popular content prediction algorithm is proposed to predict the popular contents for each SBS based on the trained AAE model. Finally, we construct a MADRL framework and then adopt the MADDPG algorithm to decide where the predicted popular contents are collaboratively cached among SBSs.

\subsection{Elastic FL Algorithm}
\begin{figure*}
	\center
	\includegraphics[scale=1]{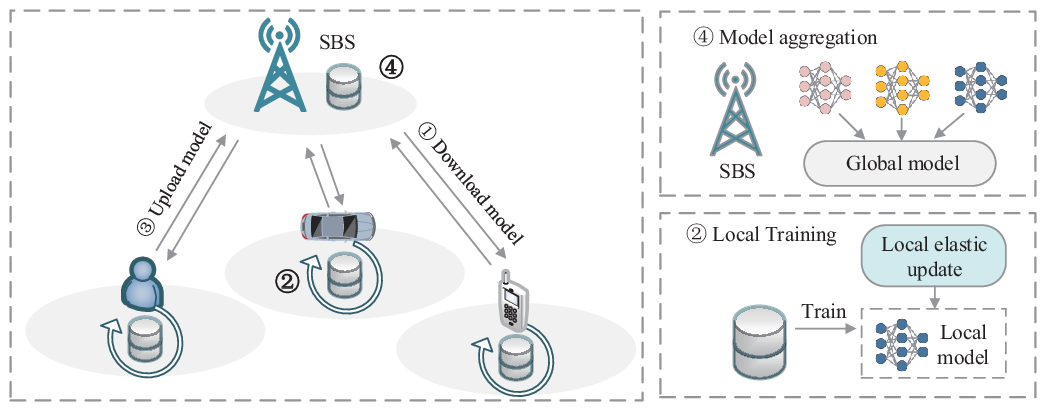}
	\caption{Elastic federated learning}
	\label{fig2}
	\vspace{-0.5cm}
\end{figure*}

As illustrated in Fig. \ref{fig2}, elastic FL algorithm is executed through $R_{max}$ rounds iteratively\cite{QiongXiaobo,Dunxing}. Within each round, the elastic FL algorithm consists of the following four steps.

\subsubsection{Download Model}
\begin{figure*}
	\center
	\includegraphics[scale=0.7]{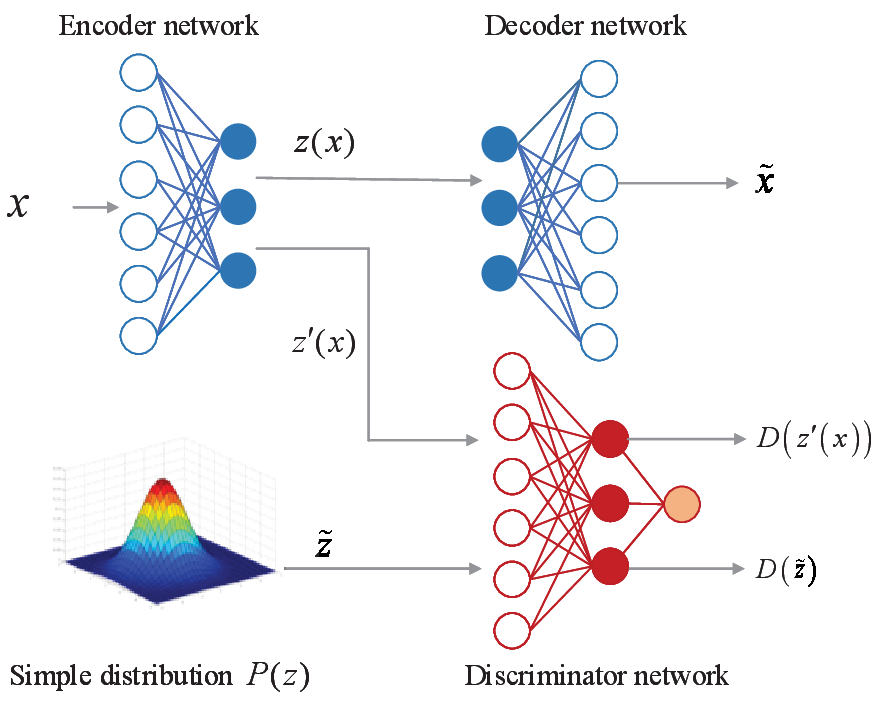}
	\caption{AAE model}
	\label{fig3}
	\vspace{-0.5cm}
\end{figure*}

Each local SBS first generates the global AAE model in this step. Let $\omega^r$ be the parameters of the global AAE model in round $r$. The global AAE model consists of the AEN and GAN. Fig. \ref{fig3} illustrates the framework of the AAE model. The upper layer of the AAE is an AEN which includes an encoder network and a decoder network. The bottom layer is a GAN which consists of a generator network and a discriminator network, where the encoder network of the AEN acts as the generator network. Let $\omega_e^r$, $\omega_d^r$ and $\omega_{d'}^r$ be the parameters of the encoder network, decoder network and discriminator network in the round $r$, respectively, thus $\omega^r = \left\{ \omega_e^r, \omega_d^r, \omega_{d'}^r \right\}$.
For the first round, the local SBS initializes its own global AAE model $\omega^0$ and for the subsequent round $r$ the local SBS updates the global AAE model $\omega^r$ at the end of round $r-1$.

\subsubsection{Local Training}
Next each UE $i$ in the coverage of local SBS downloads the global AAE model $\omega^r$ and updates the model based on its local data. If the UE directly uses the global model $\omega^r$ as the initial model for updating its own local model, it would result in the waste of the local model trained in the last round and the loss of individual UE features. Hence, we propose the elastic FL algorithm which takes into account both the local model and global model. Specifically, similar to \cite{3053055}, we first define the following weight distance formula to focus on the correlation between the local  model of UE $i$ trained in round $r-1$, i.e., $\omega_i^{r-1}$, and $\omega^{r}$, i.e.,

\begin{equation}
\operatorname{dis}\left(\omega_i^{r-1}, \omega^{r}\right)=\frac{\left\|\omega_i^{r-1}-\omega^{r}\right\|}{\left\|\omega^{r}\right\|}.
\label{eq12}
\end{equation}
As the AAE model consists of multiple layers of neural network, we introduce $\alpha_i^r$ as an elastic parameter, which denotes the gap between $\omega_i^{r-1}$ and $\omega^r$ of all layers in round $r$. Thus $\alpha_i^r$ is calculated as follows

\begin{equation}
\alpha_i^r=\frac{1}{|L|n_b} \sum_{l \in L} \operatorname{dis}\left(\omega_{i, l}^{r-1}, \omega_l^r\right),
\label{eq13}
\end{equation}
where $\omega_{i, l}^{r-1}$ is the $l$-th layer of $\omega_i^{r-1}$ and $\omega_l^r$ is the $l$-th layer of $\omega^r$. $L$ represents all the layers of the AAE model and $|L|$ denotes the number of layers. $n_b$ is the number of the UE in the SBS $b$. Then the initial local model of UE $i$ in round $r$ is calculated as
\begin{equation}
\omega_i^{r} \leftarrow \alpha_i^r \omega^r+\left(1-\alpha_i^r\right) \omega_i^{r-1},
\label{eq14}
\end{equation}
where $\alpha_i^r$ dynamically adjusts the influence of the global model on the local model, allowing the local model to have personalized features.

Then each UE updates $\omega_i^{r}$ iteratively through local training based on  its local training data.
During each iteration $k$, UE $i$ conducts a random sampling process to select training data $n_{i,k}^r$ from the overall training set. Subsequently, the training data $n_{i,k}^r$ is utilized to train the local AAE model of UE $i$.
The training process consists of the reconstruction stage and the adversarial regularization stage. The AAE model is trained alternatively between these two stages. 

The reconstruction stage which mainly focuses on training the AEN is first executed.
 Specifically, the encoder network first takes each original training data $x$ from $n_{i,k}^r$ and transforms $x$ by mapping $x$ to the network. This process yields the hidden feature representation of $x$, denoted as $z(x)$. The transformation is expressed as $z(x) = q\left(W_{i,e,k}^r x + b_{i,e,k}^r\right)$, here $W_{i,e,k}^r$ and $b_{i,e,k}^r$ are the weight matrix and bias vector of the encoder network of UE $i$ in iteration $k$ of round $r$, respectively. To make the paper clearer, we simplify $W_{i,e,k}^r$ and $b_{i,e,k}^r$ to $W_{e,k}$ and $b_{e,k}$, respectively. We also apply similar simplifications to symbols that will appear later.

 Then the decoder network calculates the reconstructed input $\tilde{x}$, i.e., $\tilde{x}=p\left(W_{d,k} z+b_{d,k}\right)$, where $W_{d,k}$ and $b_{d,k}$ are the weight matrix and bias vector of the decoder network in iteration $k$, respectively.  $p(\cdot)$ and $q(\cdot)$ are non-linear and logically-activated functions. Afterwards, the local loss function of each data $x$ for the encoder network and the decoder network is calculated as
\begin{equation}
l(\omega_{e,k}, \omega_{d,k};x)=(x-\tilde{x})^2,
\label{eq10_0}
\end{equation} where $\omega_{e,k}$ represents the parameters of the encoder network in iteration $k$, which includes $W_{e,k}$ and $b_{e,k}$, i.e., $\omega_{e,k} = \left\{W_{e,k}, b_{e,k} \right\}$.  $\omega_{d,k}$ represents the parameters of the decoder network in iteration $k$, which includes $W_{d,k}$ and $b_{d,k}$, i.e., $\omega_{d,k} = \left\{W_{d,k}, b_{d,k} \right\}$. Then the loss function of all the data in $n_{i,k}^r$ for the encoder network and decoder network is calculated as
\begin{equation}
f(\omega_{e,k}, \omega_{d,k})=\frac{1}{d_i} \sum_{x  \in n_{i,k}^r } l(\omega_{e,k}, \omega_{d,k};x),
\label{eq10}
\end{equation} where $d_i$ is the size of $n_{i,k}^r$. Let $\nabla f\left(\omega_{e,k}, \omega_{d,k}\right)$ be the gradient of $f(\omega_{e,k}, \omega_{d,k})$. Then the parameters of the encoder network and decoder network in the reconstruction stage are updated respectively as
\begin{equation}
\omega_{e,k} \leftarrow \omega_{e,k}-\eta \nabla f\left(\omega_{e,k}, \omega_{d,k}\right),
\label{eq10_2}
\end{equation}
\begin{equation}
\omega_{d,k+1} \leftarrow \omega_{d,k}-\eta \nabla f_{k}\left(\omega_{e,k}, \omega_{d,k}\right),
\label{eq10_3}
\end{equation} where $\eta$ is the local learning rate. The reconstruction stage is finished for iteration $k$. It's important to note that the encoder network will be updated again in the subsequent adversarial regularization stage. Therefore, in Eq. \eqref{eq10_2} we still adopt the notation $\omega_{e,k}$ to represent the updated parameters of the encoder network after the reconstruction stage.

 Afterwards, the adversarial regularization stage is executed, where the generator network and discriminator network will be alternately optimized. Specifically, the training of the discriminator network is first performed. The encoder network first maps $x$ to obtain a vector, i.e., $z'(x)=q\left(W_{e,k} x+b_{e,k}\right)$. It is important to note that $W_{e,k}$ and $b_{e,k}$ have been updated in the previous reconstruction stage.  
 Then $z'(x)$ will be input into the discriminator network to get its confidence $D(z'(x))$, i.e., $D(z'(x))=d\left(W_{d',k}x+b_{d',k}\right)$, where $W_{d',k}$ and $b_{d',k}$ are the weight matrix and bias vector of the discriminator network in iteration $k$, respectively. $d(\cdot)$ is also the non-linear and logically-activated function. At the same time, a vector $\tilde{z}$ is sampled from a simple predefined distribution $P(z)$. Similarly, $\tilde{z}$ is also input to the discriminator network to obtain its confidence $D(\tilde{z})$.  The objective of the discriminator network is to maximize its ability to correctly distinguish $z'(x)$ and $\tilde{z}$. Therefore, the loss function of data $x$ for the discriminator network can be calculated as,
\begin{equation}
L(\omega_{d',k};x) = -E[log D(\tilde{z})] - E[log(1 - D(z'(x)))],
\label{eq10_4}
\end{equation} where $\omega_{d',k}$ represents the parameters of the discriminator network in iteration $k$, which includes $W_{d',k}$ and $b_{d',k}$, i.e., $\omega_{d',k} = \left\{W_{d',k}, b_{d',k} \right\}$. Thus the loss function of all the data in $n_{i,k}^r$ for discriminator network is calculated as
\begin{equation}
F(\omega_{d',k})=\frac{1}{d_i} \sum_{x \in n_{i,k}^r} L(\omega_{d',k};x).
\label{eq10_5}
\end{equation} Let $\nabla F\left(\omega_{d',k}\right)$ be the gradient of $F(\omega_{d',k})$.  Then the parameters of the discriminator network are updated as
\begin{equation}
\omega_{d',k+1} \leftarrow \omega_{d',k}-\eta \nabla F_{d',k}\left(\omega_{d',k}\right).
\label{eq10_6}
\end{equation}

Then the parameters of encoder network is updated. The loss function of data $x$ for the encoder network is defined as
\begin{equation}
L(\omega_{e,k};x) = - E[log(D(z'(x)))],
\label{eq10_7}
\end{equation} 

The loss function of all the data for the encoder network is calculated as

\begin{equation}
F(\omega_{e,k})=\frac{1}{d_i} \sum_{x \in n_{i,k}^r} L(\omega_{e,k};x).
\label{eq10_8}
\end{equation} Let $\nabla F\left(\omega_{e,k}\right)$ be the gradient of $F(\omega_{e,k})$. Then the parameters of the encoder network are updated as
\begin{equation}
\omega_{e,k+1} \leftarrow \omega_{e,k}-\eta \nabla F\left(\omega_{e,k}\right).
\label{eq10_9}
\end{equation}

Iteration $k$ is finished and UE $i$ randomly samples some training data again to start the next iteration. When the number of iterations reaches the threshold $e$, UE $i$ completes the local training.

\begin{algorithm}
	\caption{The Elastic FL Algorithm}
	\label{al1}
	Initialize $\omega^{r}$;\\
	\For{round $r$ from $1$ to $R_{max}$}
	{
		\For{UE $i=1,2,...$ in parallel}
		{
			Download global AAE model $\omega^{r}$;\\
			$\omega^{r}_{i} \leftarrow \textbf{Local Updates}(\omega^r,i)$;\\
			Upload the local AAE model $\omega^{r}_{i}$ to SBS;\\
		}
		Calculate $\omega^{r+1}$ according to Eq. \eqref{eq15};\\
		\Return $\omega_{r+1}$.
	}
	\textbf{Local Update}($\omega,i$):\\
	\textbf{Input:} $\omega^r$ \\
	Calculate the $\alpha_i$ according to Eq. \eqref{eq13};\\
	Update the local AAE model $\omega^{r}_i$ according to Eq. \eqref{eq14};\\
	\For{each iteration $k$ from $1$ to $e$}
	{
		Randomly samples data $n_{i,k}^r$ from the training set;\\
		\For{data $x \in n_{i,k}^r$}
		{
			Calculate the loss function of data $x$ for the encoder network and the decoder network according to Eq. \eqref{eq10_0};\\
		}
		Calculated the local loss function of all the data for the encoder network and the decoder network in iteration $k$ according to Eq. \eqref{eq10};\\
		Update the encoder network and the decoder network according to Eq. \eqref{eq10_2} and Eq. \eqref{eq10_3};\\
		
		\For{data $x \in n_{i,k}^r$}
		{
			Calculate the loss function of data $x$ for the discriminator network according to Eq. \eqref{eq10_4};\\
		}
		Calculated the local loss function of all the data for the discriminator network in iteration $k$ according to Eq. \eqref{eq10_5};\\
		Update the discriminator network according to Eq. \eqref{eq10_6};\\
		
		\For{data $x \in n_{i,k}^r$}
		{
			Calculate the loss function of data $x$ for the encoder network according to Eq. \eqref{eq10_7};\\
		}
		Calculated the local loss function of all the data for the encoder network in iteration $k$ according to Eq. \eqref{eq10_8};\\
		Update the encoder network  according to Eq. \eqref{eq10_9};\\

	}
	\Return$\omega^r_i$.\\
\end{algorithm}
\subsubsection{Upload model}
Once each UE $i$ completes the local training, then it uploads the updated local model $\omega_i^{r}$ to the local SBS, where $\omega_i^{r}$ consists of $\omega_{i,e}^{r}$, $\omega_{i,d}^{r}$ and $\omega_{i,d'}^{r}$, i.e., $\omega_i^{r} = \left\{\omega_{i,e}^{r}, \omega_{i,d}^{r}, \omega_{i,d'}^{r} \right\}$, here $\omega_{i, e}^r$, $\omega_{i,d}^r$ and $\omega_{i,d'}^r$ be the parameters of the encoder network, decoder network and discriminator network of UE $i$ in the round $r$, respectively.

\subsubsection{Weight aggregation}
After all UEs within the coverage of the SBS upload their AAE models, the SBS generates a new global AAE model $\omega^{r+1}$ by calculating a weighted sum of all received local AAE models,
\begin{equation}
\omega^{r+1}=\omega^r-\eta \sum_{i=1} \frac{d_i}{d} \omega_i^{r},
\label{eq15}
\end{equation}
where $d$ is the size of the total data 	for all UEs within the SBS coverage.
So far, the elastic FL training for round $r$ has been completed and the SBS has obtained an updated global AAE model $\omega^{r+1}$, which is used for the training in the next round. When the number of rounds reaches $R_{max}$, the elastic FL training is finished and each SBS obtains the trained AAE model. For ease of understanding, Algorithm \ref{al1} illustrates the process of the elastic FL algorithm.
Then each SBS sends the trained AAE model to all UEs in its coverage for content popularity prediction.

\subsection{Popular Content Prediction}
\label{content popularity}
Next we propose a popular content prediction algorithm to predict the popular contents for each SBS based on the trained AAE model.
\subsubsection{Data Preprocessing}

UE $i$ abstracts a rating matrix $\boldsymbol{X}_{i}^r$ from the data in the test set. The matrix's first dimension represents users' IDs, while the second dimension is users' ratings for all contents.
Subsequently, each rating matrix $\boldsymbol{X}_{i}^r$ is fed into the updated AAE model, resulting in a reconstructed rating matrix $\boldsymbol{\tilde{X}}_{i}^r$. This reconstructed matrix is capable of extracting the hidden characteristics of the data and can be employed to approximate $\boldsymbol{X}_{i}^r$.
Then, each UE also abstracts a matrix of personal information from the test set. This matrix has the first dimension representing user IDs, and the second dimension corresponding to the personal information of the users.
%
%

\subsubsection{Cosine Similarity}

Each UE $i$ calculates the count of non-zero ratings for each user $i$ in the matrix $\boldsymbol{X}_{i}^r$ and designates the users with the top $1/m$ numbers as active users. Following this, each UE merges the matrix $\boldsymbol{\tilde{X}}_{i}^r$ and the personal information matrix (the resultant matrix is denoted as $\boldsymbol{H}_{i}^r$) to determine the similarity between each active user and the rest of the users. The similarity between an active user $u$ and another user $v$ of UE $i$ is calculated according to the cosine similarity\cite{3221271}
\begin{equation}
\begin{aligned}
\operatorname{sim}_{u,v}^{r,i}=\cos \left(\boldsymbol{H}_{i}^r(u,:), \boldsymbol{H}_{i}^r(v,:)\right)\\
=\frac{\boldsymbol{H}_{i}^r(u,:) \cdot \boldsymbol{H}_{i}^r(v,:)^T}{\left\|\boldsymbol{H}_{i}^r(u,:)\right\|_{2} \times\left\|\boldsymbol{H}_{i}^r(v,:)\right\|_{2}}
\label{eq15}
\end{aligned},
\end{equation} where $\boldsymbol{H}_{i}^r(u,:)$ and $\boldsymbol{H}_{i}^r(v,:)$  represent the vectors corresponding to active user $u$ and user $v$ in the merged matrix, respectively. The terms $\left\|\boldsymbol{H}_{i}^r(u,:)\right\|_{2}$ and $\left\|\boldsymbol{H}_{i}^r(v,:)\right\|_{2}$ stand for the 2-norm (Euclidean length) of the vectors $\boldsymbol{H}_{i}^r(u,:)$ and $\boldsymbol{H}_{i}^r(b,:)$, respectively. For each active user $u$, UE $i$ selects the users with the $K$ highest similarities as the $K$ nearest neighbors of user $u$. The ratings of these $K$ nearest users can be interpreted as indicative of the preferences of user $u$.


\subsubsection{Interested Contents}

After determining the nearest users of active users in the matrix $\boldsymbol{X}_{i}^r$, the vectors of nearest users for each active user are extracted to create a matrix $\boldsymbol{H}_K$. In this matrix, the first dimension corresponds to the IDs of the nearest users for active users, while the second dimension signifies the ratings for content from these nearest users. In the matrix $\boldsymbol{H}_K$, any content with a non-zero rating from a user is considered as the interested content of that user. The number of such interested contents is then counted for each user, and this count for a specific content is referred to as the popularity of the content. Each UE $i$ selects the contents with the top $F_p$ content popularities as the predicted interested contents and uploads these to the local SBS.
%

\subsubsection{Popular Contents}
%

After all UEs within the coverage range of the local SBS $b$ upload their predicted interesting contents, SBS $b$ collects and compares them, and selects $F_p$ most interesting contents as the predicted popular contents $p_b$ and then adopts MADDPG to cache $C$ contents from $p_b$.

\subsection{Cooperative Edge Caching Scheme Based on MADRL}
\label{MADRL}
Since each SBS has limited caching capacity, it is crucial for SBSs to collaborate in caching the predicted popular contents in order to reduce the cost for fetching contents. Therefore, we proposed the MADDPG algorithm for cooperative edge caching among SBSs, deciding SBS to cache suitable predicted popular contents, thus the cost can be optimized.
Next, the MADRL framework is first formulated, which is the basis of the MADDPG algorithm. Then, the MADDPG algorithm will be introduced.
\subsubsection{MADRL Framework}
The MADRL framework includes agents, states, actions, rewards and policies\cite{3322881,12301244,Wu20243370196}. In our cooperative edge caching, each SBS $b$ can be viewed as an agent and makes its action based on the its local state. Then we will define states, actions, rewards and policies at time slot $t$ as follows.
\paragraph{States}
At time slot $t$, the local state of SBS $b$ is defined as
\begin{equation}
\boldsymbol{s_b^t}=\left\{c_b^t, p_b^t\right\},
\label{eq1}
\end{equation}
where $p_b^t$ represents the predicted popular contents within the coverage area of SBS $b$, $c_b^t$ represents the current contents cached by SBS $b$. It should be noted that the size of $p_b^t$ is $F_p$ while the size of $c_b^t$ is $C$, and $C < F_p$. The global state is defined as
\begin{equation}
\boldsymbol{s^t}=\left\{s_1^t, \ldots, s_b^t, \ldots, s_B^t\right\}.
\label{eq2}
\end{equation}

\paragraph{Actions}
Once SBS $b$ predicts new popular contents $p_b^t$ at time slot $t$, it needs to select an appropriate contents from $p_b^t$ to replace part of the contents in $c_b^t$.
Therefore, at time slot $t$, the action of SBS $b$ is defined as
\begin{equation}
\boldsymbol{a_b^t}=\left\{a_{b, 1}^t, \ldots, a_{b, f}^t, \ldots, a_{b, F_p}^t\right\},
\label{eq3}
\end{equation}
where each $a_{b, f}^t$ is a binary variable indicating if the $f$-th content in $p_b$ is stored or not. $0$ stands for that the $f$-th popular content is not stored in SBS $b$, while $1$ indicates that it will be cached. Let $r_{b, e}^t$ be the number of $1$ in \boldsymbol{$a_{b}^t$}, thus the number of $0$ is $F_p - r_{b, e}^t$.
The global action is defined as
\begin{equation}
\boldsymbol{a^t}=\left\{a_1^t, \ldots, a_b^t, \ldots, a_B^t\right\}.
\label{eq4}
\end{equation}

\paragraph{Reward function}
Compared to UEs fetching contents from CS, it can significantly reduce costs by using the cooperative edge caching system. Thus we define the reward as the saved cost. Next, we will first introduce the cost which UEs fetch contents from the cooperative edge caching system, then introduce the cost which UEs fetch all contents only from CS, and the difference between these two costs is the saved cost, which is our defined reward.

In the cooperative edge caching system, a UE can fetch requested contents directly from the local SBS, CS or adjacent SBSs. The three optional methods of obtaining the requested contents usually correspond to different costs.

The UE can fetch the requested contents from the local SBS $b$. Let $\alpha$ denote the cost of fetching the contents from local SBS $b$. Assume that UEs fetch $r_b^t$ contents from local SBS $b$ at the time slot $t$, thus the cost is $\alpha r_b^t$.

The UE can indirectly fetch the requested contents from adjacent SBSs. Let $\beta$ denote the cost that the UEs within the coverage of local SBS $b$ fetch the contents from adjacent SBSs. Since the delivery of contents between SBSs will consume backhaul resources, the cost of fetching contents from adjacent SBSs is higher than that of fetching the contents from local SBS $b$, i.e., $\beta>\alpha$.
Assume that the UEs in the coverage of local SBS $b$ fetch $r_{b, n}^t$ contents from adjacent SBSs. In this case, the cost is $\beta r_{b, n}^t$.

The UE can fetch the requested content from CS. Let $\chi$ denote the cost of fetching the contents from CS. The cost of fetching contents from CS is much higher than that of fetching contents from SBSs owing to the resource consumption of backhaul and core network, i.e., $\chi>\beta$ and $\chi>\alpha$. Assuming that the UEs in the coverage of local SBS $b$ fetch $r_{b, c}^t$ contents from CS at time slot $t$. In this case, the cost is $\chi r_{b, c}^t$.

In addition, the contents replaced by SBS $b$ are all from CS, it will result in the additional costs as it requires the consumption of backhaul resources between SBS $b$ and CS. Denote $\delta$ as the cost of replacing a content. The cost of replacing caching contents is $\delta r_{b, e}^t$. Thus, the total cost of SBS $b$ is $\alpha r_b^t+\beta r_{b, n}^t+\chi r_{b, c}^t+\delta r_{b, e}^t$.

Noted that the constants $\alpha$, $\beta$, $\chi$ and $\delta$ can be represented with actual physical significance, that is, these values can be explained as a general expression of the benefits from content caching, such as energy saving, expense reduction and so on\cite{Chen3044298,2713384,9417383}.

Edge caching in the cooperative edge caching system can significantly reduce costs. Without edge caching, all UEs would have to fetch contents from CS, which leads to a great cost, i.e., $\chi\left(r_b^t+r_{b, n}^t+r_{b, c}^t\right)$, thus the reward of SBS $b$ at time slot $t$ can be defined as the saved cost, i.e.,
\begin{equation}
\begin{aligned}
& R_b^t=\chi\left(r_b^t+r_{b, n}^t+r_{b, c}^t\right)\\
&-\left(\alpha r_b^t+\beta r_{b, n}^t+\chi r_{b, c}^t+\delta r_{b, e}^t\right) \\
& =(\chi-\alpha) r_b^t+(\chi-\beta) r_{b, n}^t-\delta r_{b, e}^t
\end{aligned}.
\label{eq5}
\end{equation} Note that the higher the cost savings, i.e., the larger $R_b^t$, will cause more effective edge caching. We use  $\boldsymbol{R_L^t}=\left[R_1^t, \ldots, R_b^t, \ldots, R_B^t\right]$ to denote all SBSs' reward.
The global reward at time slot $t$ is calculated as
\begin{equation}
R^t=\frac{1}{B} \sum_{b=1}^B R_b^t.
\label{eq6}
\end{equation}

Let $\boldsymbol{\pi^t}=\left\{\pi_1^t, \ldots, \pi_b^t, \ldots, \pi_B^t\right\}$ denotes the caching policies at time slot $t$. $\boldsymbol{\pi^t}$ will map state \boldsymbol{$s^t$} to action $\boldsymbol{a^t$}, i.e., $\boldsymbol{a^t} = \boldsymbol{\pi(s^t)}$.
The expected long-term discounted reward of MADRL is calculated as
\begin{equation}
J(\boldsymbol{\pi^t})=\mathrm{E}\left[\sum_{t=0}^{\infty} \gamma^t R^t\right].
\label{eq7}
\end{equation}
where $\gamma \in(0,1)$ is the discount factor. Our cooperative edge caching problem can be formulated as a multi-agent decision problem to maximize the expected long-term discounted reward to get the optimal policies $\boldsymbol{\pi^*}$, which is calculated as
\begin{equation}
\boldsymbol{\pi^*}=\arg \min J(\boldsymbol{\pi}).
\label{eq8}
\end{equation}

\subsubsection{MADDPG algorithm}
The MADDPG algorithm is proposed for the cooperative edge caching. Next we first introduce the training stage of MADDPG algorithm to obtain the optimal policies, then describe the testing stage to test performance based on the optimal policies.

The MADDPG algorithm is based on the multi-agent actor-critic framework \cite{3259688}.
Each agent $b$ has a local actor network, a local critic network, a local target actor network and a local target critic network. The local actor network of agent $b$ outputs the policy based on the local state, then the action can be obtained according to the policy. The local critic network of agent $b$ is used to evaluate the action according to the local reward. Two local target networks can ensure the stability of the algorithm, whose network structure is same as that of local actor network and local critic network, respectively. All agents share two centralized global critic networks and two global target critic networks, all of them are located at CS.
Two global critic networks provide a global perspective based on the global states and actions of all agents, enabling each agent to understand and adapt to the global environment. Two global target critic networks which have the same network structure as the global critic networks, respectively, are also used to ensure the stability.

Let $\theta_b$ and $\phi_b$ are the parameters of the local actor network and local critic network of agent $b$, respectively, while $\theta_b^{\prime}$ and $\phi_b^{\prime}$ are the parameters of the local target actor network and local target critic network of agent $b$, respectively. Let $\varphi_1$, $\varphi_2$, $\varphi_1^{\prime}$ and $\varphi_2^{\prime}$ be the parameters of two global critic networks and two global target critic networks, respectively.
The detailed training stage for MADDPG algorithm is further introduced, and Algorithm \ref{al2} describes the specific process of MADDPG training stage.

\begin{algorithm}
  \caption{Training Stage for the MADDPG Framework}
  \label{al2}
  \KwIn{$\gamma$, $\tau$, $\theta_b$, $\phi_b$, $\varphi_1$, $\varphi_2$}
  Randomly initialize the  $\theta_b$, $\phi_b$, $\varphi_1$, $\varphi_2$\;
  Initialize target networks by $\theta'_b \leftarrow \theta_b$, $\phi'_b \leftarrow \phi_b$, $\varphi'_1 \leftarrow \varphi_1$, $\varphi'_2 \leftarrow \varphi_2$\;
  Initialize replay buffer $\mathcal{D}$\;
  \For{episode from $1$ to $R_{max}$ }
  {
	  \For{agent $b$ from 1 to $B$}
  {
	Predict the popular contents $p_b$;\\
    Receive initial observation state $s^{1}_b$\;
	}
    $s^1=\left[s_1^1, \ldots, s_b^1, \ldots, s_B^1\right]$;\\
    \For{time slot $t$ from $1$ to $T$ }
    {
	  \For{agent $b$ from 1 to $B$}
  {
		Observe $s^t_b$ and select action $a_b^t=\pi\left(s_b^t \mid \theta_b\right)$;\\
	}
	  $s^t=\left[s_1^t, \ldots, s_b^t, \ldots, s_B^t\right]$, $a^t=\left[a_1^t, \ldots, a_b^t, \ldots, a_B^t\right]$;\\
	  Receive global and local rewards, $R^t$ and $R^t_L$;\\
      Store transition $(s^t,a^t,R^t,R^t_L,s^{t+1})$ in $()()$\;
      \If {number of transition tuples in $\mathcal{D}$ is larger than $M$ }
      {
      Randomly sample a mini-batch of $M$, $(s^i,a^i,R^i,R^i_L,s')$, from $\mathcal{D}$\;
      Set $y_g^i=R^i+\gamma \min _{x=1,2} Q\left(s^{\prime i}, a^{\prime i} \mid \varphi_x^{\prime}\right)$;\\
	  Update global critics by minimizing the loss based on Eq. \eqref{eq21};\\
	  Update target parameters based on Eq. \eqref{eq22};\\
	  \For{agent $b$ from 1 to $B$}
  {
	  Set $y_b^i=R_b^i+\gamma Q\left(s_b^{\prime i}, a_b^{\prime i} \mid \phi_b^{\prime}\right)$;\\
	  Update local critic network by minimizing the loss based on Eq. \eqref{eq24};\\
	  Update the local actor network based on Eq. \eqref{eq25};\\
	  Update target local networks according to Eqs. \eqref{eq26} and \eqref{eq27}.\\
}
      }
    }
  }
\end{algorithm}

\paragraph{Training Stage}
Firstly, each agent $b$ randomly initializes its local actor network parameters $\theta_b$ and local critic network parameters $\phi_b$. CS also initializes the two global critic network parameters $\varphi_1$ and $\varphi_2$. The local target actor network parameters $\theta_b^{\prime}$ and local target critic network parameters $\phi_b^{\prime}$ of each agent $b$ are initialized as $\theta_b$ and $\phi_b$, while two target global critic network parameters $\varphi_1^{\prime}$ and $\varphi_2^{\prime}$ are initialized as $\varphi_1$ and $\varphi_2$.
The replay buffer $D$ which is located at CS is equipped with substantial storage capacity to store transition tuples in each time slot $t$, i.e., $\boldsymbol{s^t}$, $\boldsymbol{a^t}$, $R^t$, $\boldsymbol{R_L^t}$ and $\boldsymbol{s^{t+1}}$. Replay buffer $D$ can facilitate the efficient sharing of experience across the agents.

Next, the algorithm executes for $R_{max}$ episodes. In the first episode, each SBS $b$ obtains $F_p$ popular contents $p_b^1$ from predicted interested contents uploaded by UEs within its own coverage, and randomly selects $C$ contents for caching. Thus each SBS $b$ can obtain the initial caching contents $c_b^1$. Based on $c_b^1$ and $p_b^1$, each SBS $b$ can obtain its initial local state $\boldsymbol{s_b^1}$ according to Eq. \eqref{eq1}, and the initial global state can also be further obtained according to Eq. \eqref{eq2}, i.e., $\boldsymbol{s^1}=\left[s_1^1, \ldots, s_b^1, \ldots, s_B^1\right]$.

Then algorithm is iteratively executed from slot $1$ to slot $T$. In the first slot,  each SBS $b$ inputs its local state $s_b^1$ to its local actor network and gets its policy $\pi_b^1$, then the local action $a_b^1$ can be obtained based on $\pi_b^1$. Thus the global action can be obtained according to Eq. \eqref{eq4}, i.e., $\boldsymbol{a^1}=\left[a_1^1, \ldots, a_b^1, \ldots, a_B^1\right]$. After each SBS $b$ takes the local action $a_b^1$, it can get its local reward $R_b^1$ according to Eq. \eqref{eq5}. Thus all agents' rewards can be obtained, i,e., $\boldsymbol{R_L^1}=\left[R_1^1, \ldots, R_b^1, \ldots, R_B^1\right]$. The global reward $R^1$ can also be obtained according to Eq. \eqref{eq6}. In addition, each SBS $b$ enters the next local state $s_b^2$ after taking the local action, and the next global state $s^2$ can be further obtained, i.e., $\boldsymbol{s^2}=\left[s_1^2, \ldots, s_b^2, \ldots, s_B^2\right]$.

Then the transition tuple $\left(\boldsymbol{s^1}, \boldsymbol{a^1}, R^1, \boldsymbol{R_L^1}, \boldsymbol{s^2}\right)$ is stored in the replay buffer $D$. When the number of transition tuples stored in the replay buffer $D$ is less than $M$, the algorithm starts the next slot.

\begin{figure*}
\center
\includegraphics[scale=0.23]{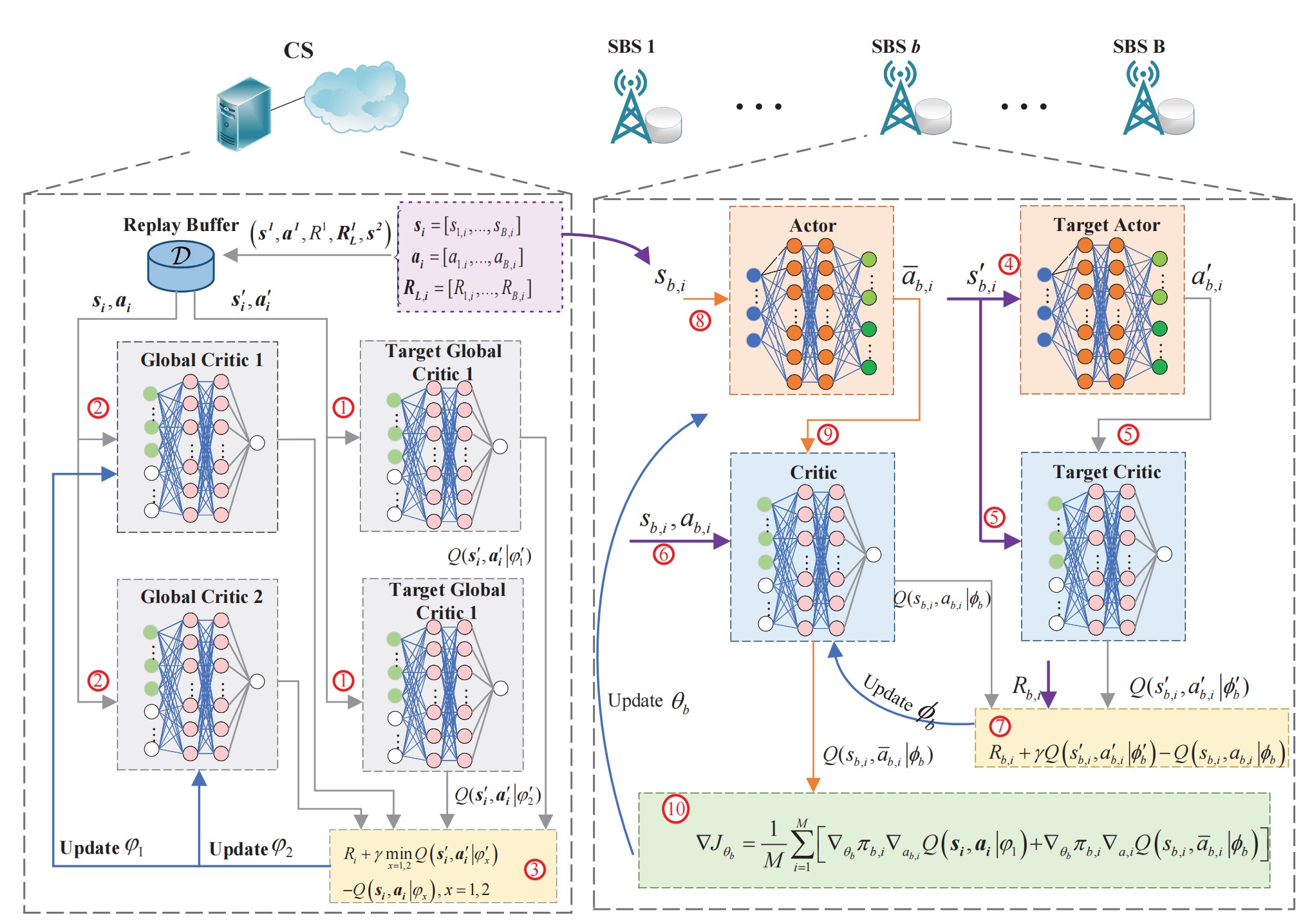}
\caption{MADDPG algorithm flow chart of the training stage}
\label{fig4}
\vspace{-0.5cm}
\end{figure*}
When the number of stored transition tuples is greater than $M$ in time slot $t$, CS first randomly samples $M$ transition tuples from the replay buffer $D$ to form a mini-batch. For the sake of simplicity, let $\left(\boldsymbol{s_i}, \boldsymbol{a_i}, R_i, \boldsymbol{R_{L,i}}, \boldsymbol{s_i^{\prime}}\right)(i=1,2, \ldots, M)$ represents the $i$-th transition tuple in the mini-batch. For the transition tuple $i$, SBS $b$ first inputs its local state $s_{b,i}^{\prime}$ into the local target local actor network and gets its policy $\pi^{\prime}_{b,i}$, then the local action $a_{b,i}^{\prime}$ can be obtained based on $\pi^{\prime}_{b,i}$. All agents' actions $\boldsymbol{a_i^{\prime}}=\left[a_{1,i}^{\prime}, \ldots, a_{b,i}^{\prime}, \ldots, a_{B,i}^{\prime}\right]$ can be obtained. Then CS inputs $\boldsymbol{a_i}^{\prime}$ and $\boldsymbol{s_i^{\prime}}$ into two target global critic networks, and outputs two target action-state values $Q\left(\boldsymbol{s_i^{\prime}}, \boldsymbol{a_i^{\prime}} \mid \varphi_1^{\prime}\right)$ and $Q\left(\boldsymbol{s_i^{\prime}}, \boldsymbol{a_i^{\prime}} \mid \varphi_2^{\prime}\right)$, respectively. This process is shown as step 1 in Fig. 4. The target global value can be calculated as
\begin{equation}
y_{g,i}=R_i+\gamma \min _{x=1,2} Q\left(\boldsymbol{s_i^{\prime}}, \boldsymbol{a_i^{\prime}} \mid \varphi_x^{\prime}\right).
\label{eq20}
\end{equation}

Next CS inputs $\boldsymbol{a_i}$ and $\boldsymbol{s_i}$ into two global critic networks, and outputs two action-state values $Q\left(\boldsymbol{s_i}, \boldsymbol{a_i} \mid \varphi_1\right)$ and $Q\left(\boldsymbol{s_i}, \boldsymbol{a_i} \mid \varphi_2\right)$, respectively. This process is shown as step 2 in Fig. 4. The loss function of each global critic network is calculated as

\begin{equation}
L\left(\varphi_x\right)=\frac{1}{M} \sum_{i=1}^M\left[y_{g,i}-Q\left(\boldsymbol{s_i}, \boldsymbol{a_i} \mid \varphi_x\right)\right]^2, x=1,2,
\label{eq21}
\end{equation}

The above equation is shown as step 3 in Fig. 4. The gradient of the each global critic network can be calculated as $\nabla L\left(\varphi_x\right), x=1,2$.

Then each SBS $b$ trains its own local critic and local actor networks. Each SBS $b$ first inputs its local state $s_{b,i}^{\prime}$ into its local target actor network and obtains the local action $a_{b,i}^{\prime }$. This process is shown as step 4 in Fig. 4. Next each SBS $b$ inputs $s_{b,i}^{\prime}$ and $a_{b,i}^{\prime }$ into its local target critic network, and gets the local target action-state value $Q\left(s_{b,i}^{\prime}, a_{b,i}^{\prime  } \mid \phi_b^{\prime}\right)$. This process is shown as step 5 in Fig. 4. Thus the target value of the local critic network can be calculated as
\begin{equation}
y_{b,i}=R_{b,i}+\gamma Q\left(s_{b,i}^{\prime}, a_{b,i}^{\prime  } \mid \phi_b^{\prime}\right).
\label{eq23}
\end{equation}
Then each SBS $b$ inputs $s_{b,i}$ and $a_{b,i}$ into its local critic network, and gets the local action-state value $Q\left(s_{b,i}, a_{b,i}\mid \phi_b\right)$. This process is shown as step 6 in Fig. 4. The local critic loss function of SBS $b$ can be calculated as
\begin{equation}
L\left(\phi_b\right)=\frac{1}{M} \sum_{i=1}^M\left[y_{b,i} -Q\left(s_{b,i}, a_{b,i} \mid \phi_b\right)\right]^2.
\label{eq24}
\end{equation}
The above equation is shown as step 7 in Fig. 4. Then the gradient of the local critic network is calculated as $\nabla L\left(\phi_b\right)$. Then each SBS $b$ inputs $s_{b,i}$ into its local actor network and gets its policy $\bar{\pi}_{b,i}$, then the local action $\bar{a}_{b,i}$ can be obtained based on $\bar{\pi}_{b,i}$. Next each SBS $b$ inputs $s_{b,i}$ and $\bar{a}_{b,i}$ into its local target critic network, and gets the local target action-state value $Q\left(s_{b,i}, \bar{a}_{b,i} \mid \phi_b \right)$. These processes is shown as steps 8 and 9 in Fig. 4. The gradient of the local actor network of SBS $b$ is calculated as
\begin{equation}
\begin{aligned}
& \nabla J_{\theta_b}=\frac{1}{M} \sum_{i=1}^M\left[\nabla_{\theta_b} \pi_{b,i} \nabla_{a_{b,i}} Q\left(\boldsymbol{s_i}, \boldsymbol{a_i} \mid \varphi_1\right)\right. \\
& \qquad \qquad \qquad \left.+\nabla_{\theta_b} \pi_{b,i}  \nabla_{a_{b,i}} Q\left(s_{b,i}, \bar{a}_{b,i} \mid \phi_b\right)\right].
\end{aligned}
\label{eq25}
\end{equation}
The above equation is shown as step 10 in Fig. 4.
The first term in Eq. \eqref{eq25} is associated with the global critic network, similar with \cite{3259688}, we only use one action-state value from one of the global critic networks, i.e., $Q\left(\boldsymbol{s_i}, \boldsymbol{a_i} \mid \varphi_1\right)$.
The second term corresponds to each agent's local critic network.

After that $\varphi_1$,  $\varphi_2$, $\phi_b$ and $\theta_b$ are updated through gradient ascending based on $\nabla L\left(\varphi_1\right)$, $\nabla L\left(\varphi_2\right)$, $\nabla L\left(\phi_b\right)$ and $\nabla J_{\theta_b}$.

Then the parameters of the two target global critic networks, local actor network and the critic network are updated as
\begin{equation}
\varphi_x^{\prime}=\tau \varphi_x+(1-\tau) \varphi_x^{\prime}, x=1,2,
\label{eq22}
\end{equation}

\begin{equation}
\theta_b^{\prime}=\tau \theta_b+(1-\tau) \theta_b^{\prime},
\label{eq26}
\end{equation}
\begin{equation}
\phi_b^{\prime}=\tau \phi_b+(1-\tau) \phi_b^{\prime}.
\label{eq27}
\end{equation} where $\tau \leq 1$ is a constant value. The update in this time slot is finished, and the above procedure is repeated for the next time slot. When the number of time slot reaches $T$, the current episode ends up. Then each SBS initialize its cached contents to start the next episode. When the number of episodes reaches $N$, the algorithm will terminate and we will get the optimal parameters $\theta_b^*$ of the local actor network of each SBS $b$. The flow chart of the MADDPG training stage is shown in Fig. \ref{fig4}.

\paragraph{Testing Stage}
The testing stage omits the critic network, target actor network, target critic network of each SBS $b$, two global critic networks and two target global critic networks, as compared to the training stage. During the testing stage, the local state $s_b^t$ of each SBS $b$ in each time slot $t$ is input into its local actor network with optimized parameters $\theta_b^*$, and the optimal action which denotes the best caching placement can be obtained. The pseudo-code of the testing stage is shown in Algorithm \ref{al3}.

\subsection{Computational Complexity Analysis}
In this section, we will analyze the computational complexity of our approach. The analysis focuses on the training stage due to the significant computation resource and time consumption in the training stage. Our methodology for analyzing the computational complexity is inspired by \cite{hunbiaozhu}.
During a round, the training process includes two processes: the elastic FL training of AAE and the training of MADDPG. Because the training process requires significant computational resources to calculate gradients and update parameters, the computational complexity mainly consists of the complexity of computing gradients and the complexity of updating parameters.

Let \(G_{E}\), \(G_{D}\), \(G_{D’}\), \(G_{A}\), and \(G_{C}\) be the computational complexity of computing gradients for the encoder network, the decoder network, the discriminator network, the actor network, and the critic network, respectively. Let \(U_{E}\), \(U_{D}\), \(U_{D’}\), \(U_{A}\), and \(G_{C}\) be the computational complexity of updating parameters for the aforementioned networks. First, we analyze the complexity of the elastic FL training of AAE, which is influenced by the number of SBSs and the number of UEs covered by each SBS. Let \(n_b\) represent the number of UEs under SBS b, and each UE needs to undergo \(e\) iterations of local training. Therefore, the complexity of the elastic FL training of AAE is \(O_{aarfl} = O\left( R_{max}\sum^{B}_{b=1} n_{b}e\left( 2G_{E}+G_{D}+G_{D^{\prime }}+2U_{E}+U_{D}+U_{D^{\prime }}\right)  \right)\).

Next, we analyze the computational complexity of MADDPG training, which is mainly influenced by the number of SBSs. Since the local target actor and local actor have the same network structure, they have the same \(G_{A}\) and \(U_{A}\). Moreover, each SBS's two local target critic networks, two local critic networks, two global critic networks, and two global target critic networks have the same network structure, so they are both \(G_{C}\) and \(U_{C}\). It should be noted that the target networks do not need to caculate gradients, and the training and updating parameters process will not be activated until the tuples stored in the replay buffer are larger than \(M\). Therefore, the computational complexity of MADDPG training is \(O_{madrl} = O((R_{max}T-M)\left( B(G_{A}+U_{E})+(2+B)(G_{C}+U_{E}\right)  ))\).

Thus, the total computational complexity is \(O = O_{aarfl} + O_{madrl}\).
\begin{algorithm}
  \caption{Testing Stage for the MADDPG Framework}
  \label{al3}
  \For{episode from $1$ to $E'$ }
  {
    Predict the popular contents $p_b^1$;\\
    Receive initial local state $s_b^1$\;
    \For{time slot $t$ from $1$ to $T$ }
    {
	\For{agent $b$ from 1 to $B$}
  {
Generate caching placement based on the optimal policy $a_b^t=\pi\left(s_b^t \mid \theta^*_b\right)$ \;
Execute action $a_{b}^t$, observe local reward $R_b^t$ and obtain new local state $s^{t+1}_b$;\
	}
Observe global $R^t$ and obtain next global state $\boldsymbol{s^{t+1}}$.
    }
  }
\end{algorithm}

\section{Simulation Results}
\label{sec6}

\begin{table}
\caption{Values of the parameters in the experiments.}
\label{tab2}
\footnotesize
\centering
\begin{tabular}{|p{1.3cm}<{\centering}|p{1.2cm}<{\centering}|p{1.3cm}<{\centering}|p{1.2cm}<{\centering}|}
  \hline
  \textbf{Parameter} &\textbf{Value} &\textbf{Parameter} &\textbf{Value}\\
  \hline
	$R_{max}$ & $1000$ & $E^{\prime}$ &$100$\\
	\hline
  $M$ &$256$ & $T$  & $100$\\
	\hline
	$\alpha$ & $1$ & $\beta$ & $30$ \\
  \hline
  $\chi$ & $100$ & $\delta$ & $100$\\
  \hline
	$\gamma$ & $0.99$ & $\eta$ & $0.01$\\
  \hline
  $\tau$ & $0.001$ & & \\
  \hline
\end{tabular}
\end{table}
In this section, simulation experiments are conducted to evaluate the performance of the proposed CEFMR scheme in the collaborative caching system.

\subsection{Settings and Dataset}

The simulation experiments in this paper are conducted based on Python 3.8 while leveraging the extensively acknowledged MovieLens 1M dataset, which encompasses nearly 100 million movie ratings given by 6040 anonymous users for 3883 movies \cite{10039430}. A part of the movies that have been rated by users will be  the requested contents for the UEs. Table \ref{tab2} lists the values of all parameters in the simulation.

\subsection{Training Stage}
\begin{figure}
\center
\includegraphics[scale=0.45]{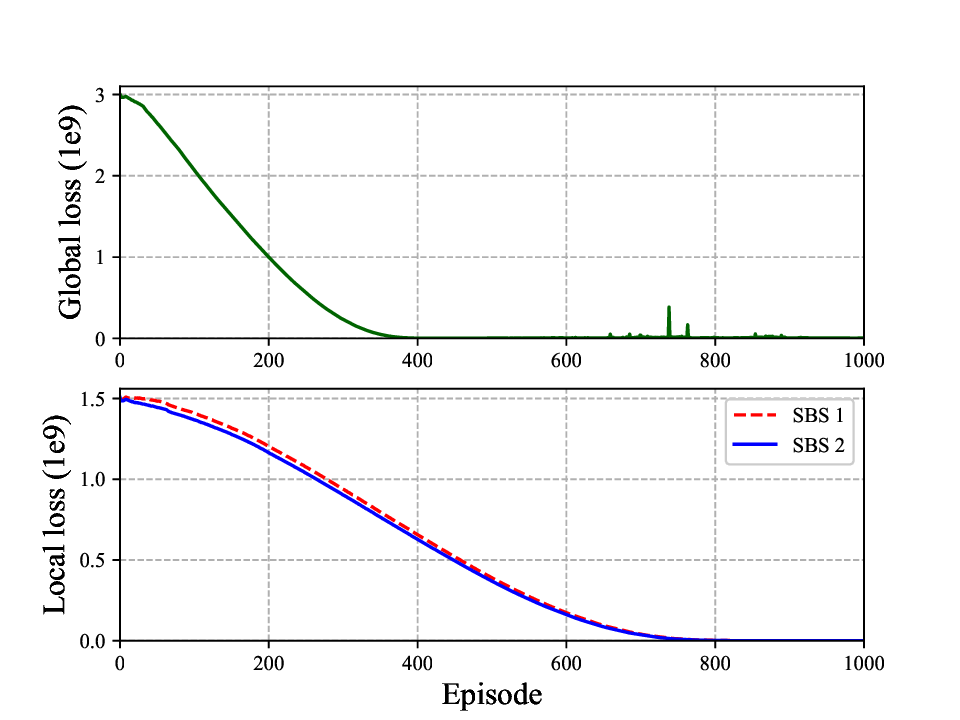}
\caption{Global and local losses during the training stage of MADDPG}
\label{fig7}
\vspace{-0.5cm}
\end{figure}

First, we present the training results of the MADDPG algorithm, where the number of SBSs is set to $2$. Fig. \ref{fig7} shows the average loss of the two global critic networks for the two SBSs, and two local losses of the local critic networks for each SBS. As can be seen from Fig. \ref{fig7}, the global loss decreases slowly until the $400th$ episode and then gradually stabilizes, while the local losses of two SBSs decrease slowly until the $750th$ episode and then gradually stabilize. The loss represents the error between the action-state value and the actual reward, thus the action-state value gradually approaches the actual reward with the progress of training the loss decreases. This means that MADDPG algorithm converges after training, and we can obtain an optimal local actor network for each SBS.
It should be noted that the number of UEs will only affect the speed of elastic FL training and does not impact MADDPG. Additionally, the primary objective of this paper is to address the cooperative caching challenges among multiple MBSs. As a result, to maintain focus and ensure a controlled environment for our analysis, we set a fixed number of UEs during the training of elastic FL. Research on dynamic UE numbers has been explored in our previous work\cite{3221271,10039430}.

\begin{figure}[h]
  \centering
  \subfigure[]{
    \begin{minipage}{7.5cm}
    \includegraphics[scale=0.45]{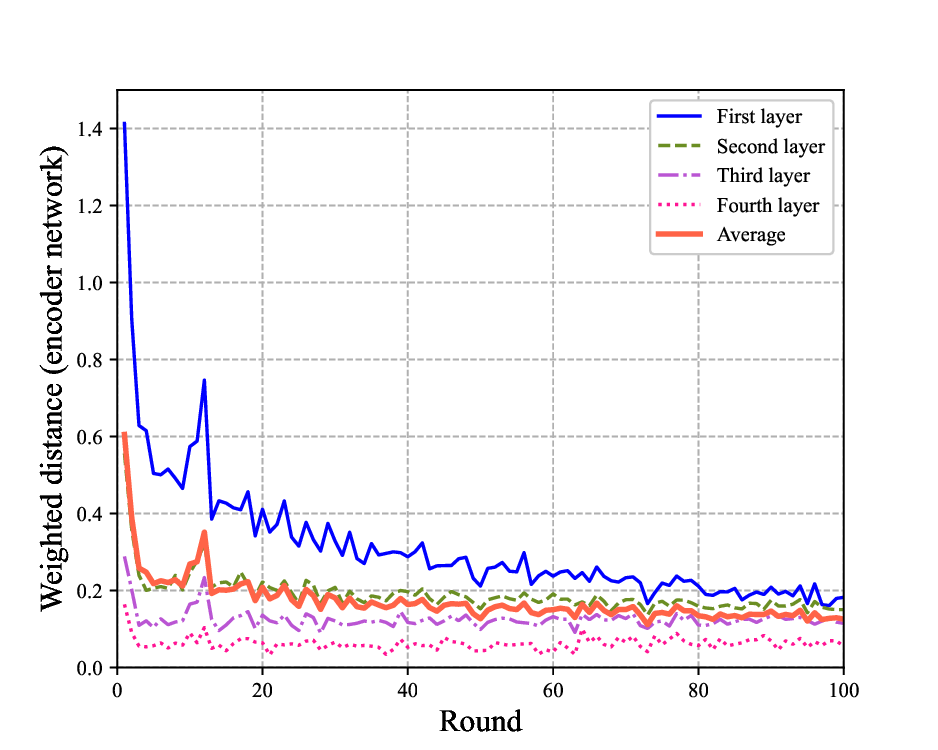}
    \end{minipage}}
  \subfigure[]{
    \begin{minipage}{7.5cm}
    \includegraphics[scale=0.45]{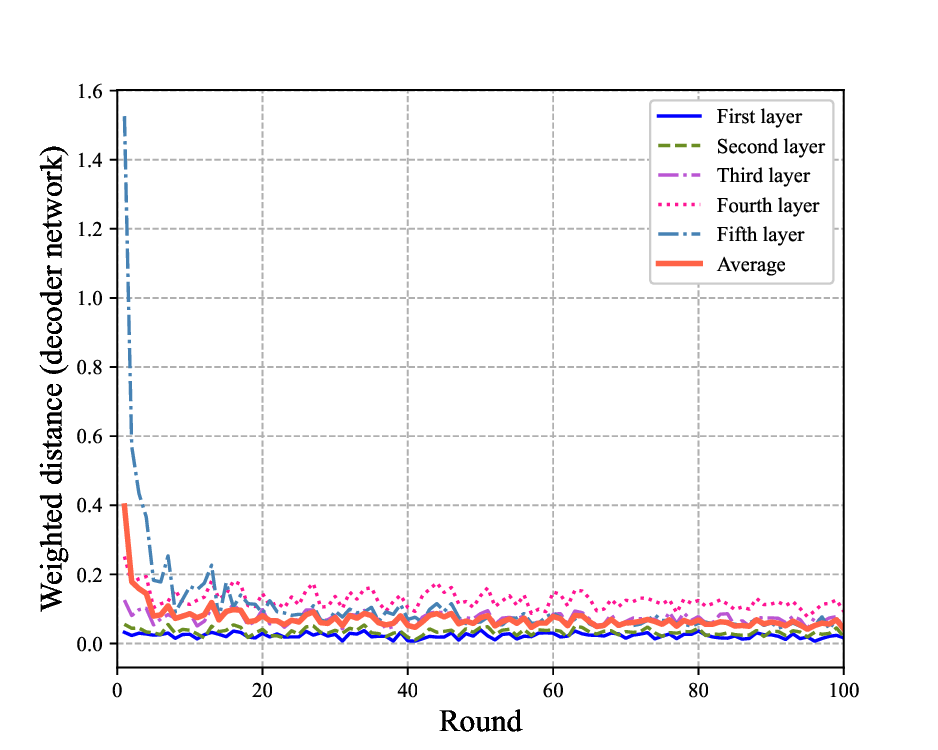}
    \end{minipage}}
  \subfigure[]{
    \begin{minipage}{7.5cm}
    \includegraphics[scale=0.45]{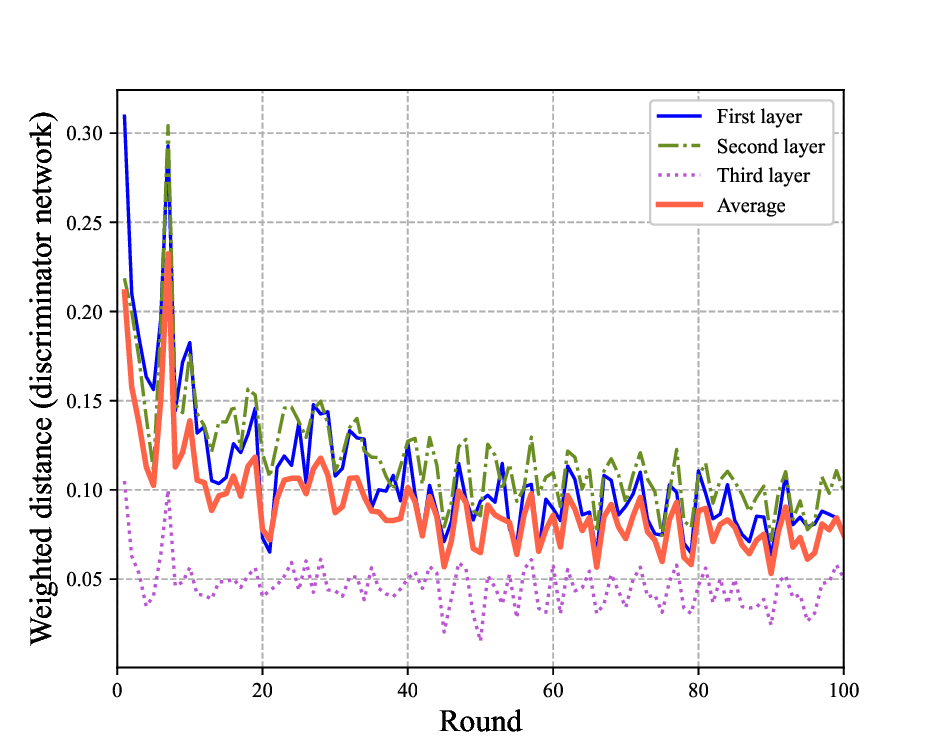}
    \end{minipage}}
    \caption{Weight distance vs round. (a) Encoder network; (b) Decoder network; (c) Discriminator network.}
    \label{fig9}
    \vspace{-0.5cm}
\end{figure}

\begin{figure}[h]
  \centering
    \subfigure[]{
    \begin{minipage}{7.5cm}
    \includegraphics[scale=0.45]{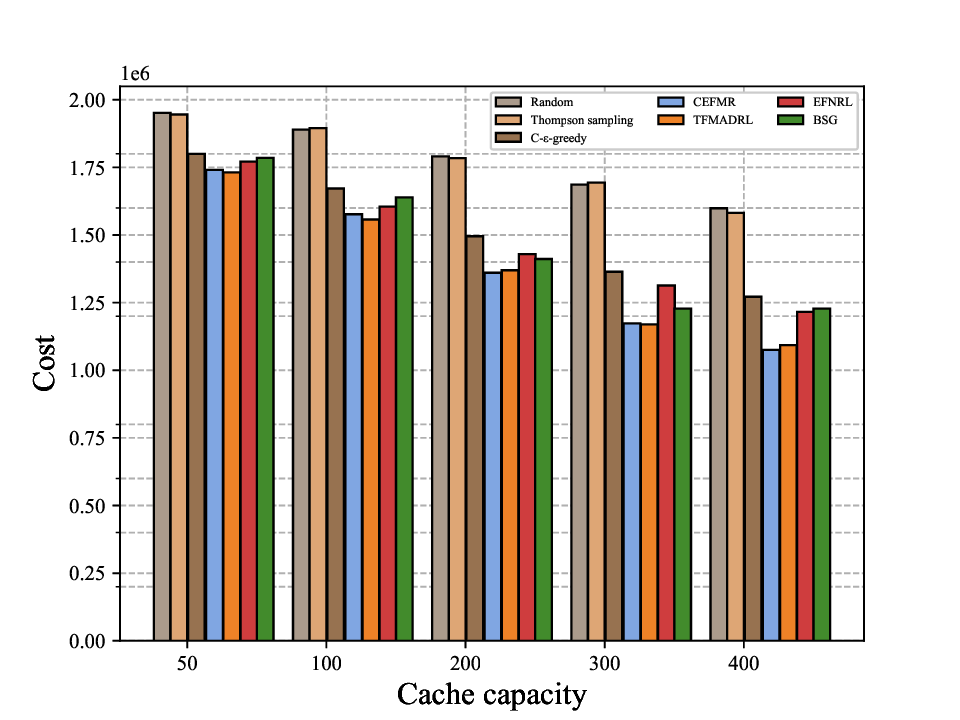}
    \end{minipage}}
  \subfigure[]{
    \begin{minipage}{7.5cm}
    \includegraphics[scale=0.45]{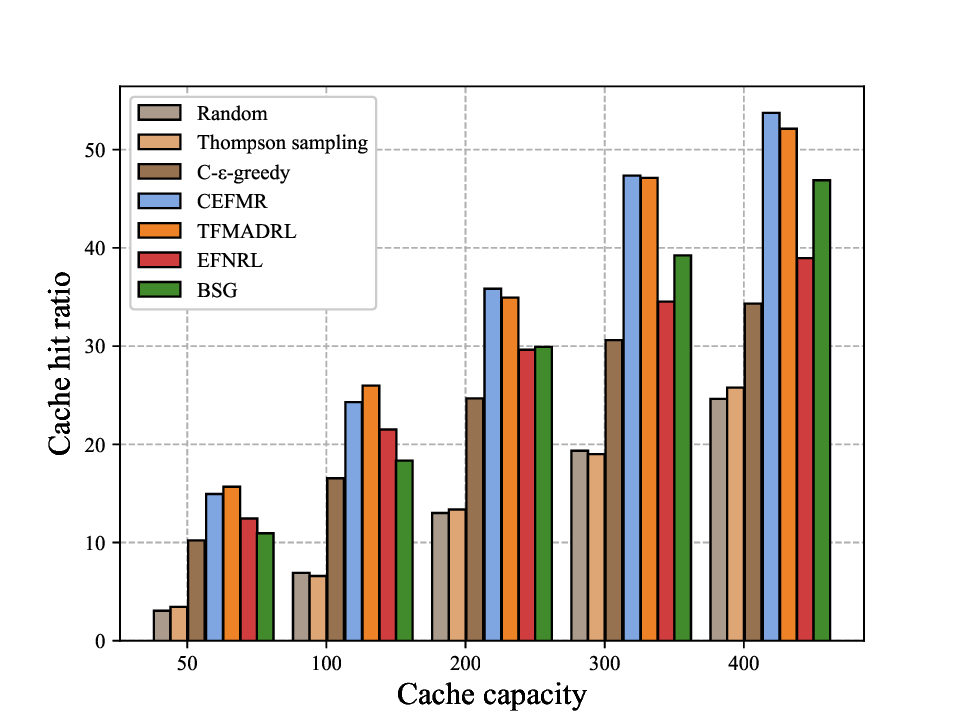}
    \end{minipage}}
    \caption{Performance vs cache capacity under different schemes. (a) Cost; (b) Cache hit ratio.}
    \label{fig5}
    \vspace{-0.5cm}
\end{figure}

\begin{figure}[h]
  \centering
  \subfigure[]{
    \begin{minipage}{7.5cm}
    \includegraphics[scale=0.45]{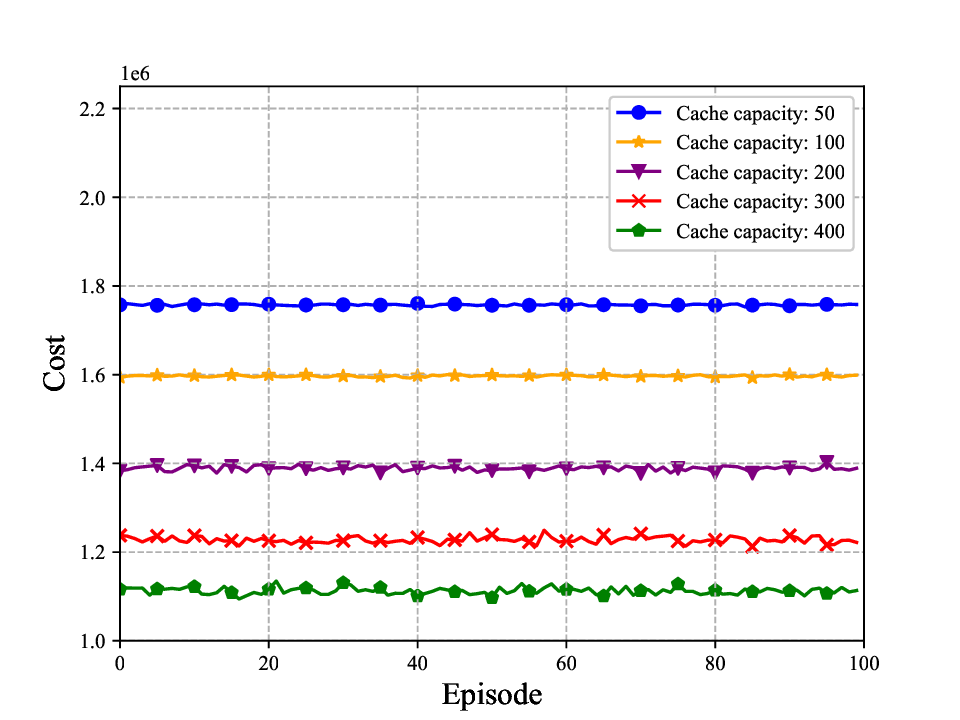}
    \end{minipage}}
  \subfigure[]{
    \begin{minipage}{7.5cm}
    \includegraphics[scale=0.45]{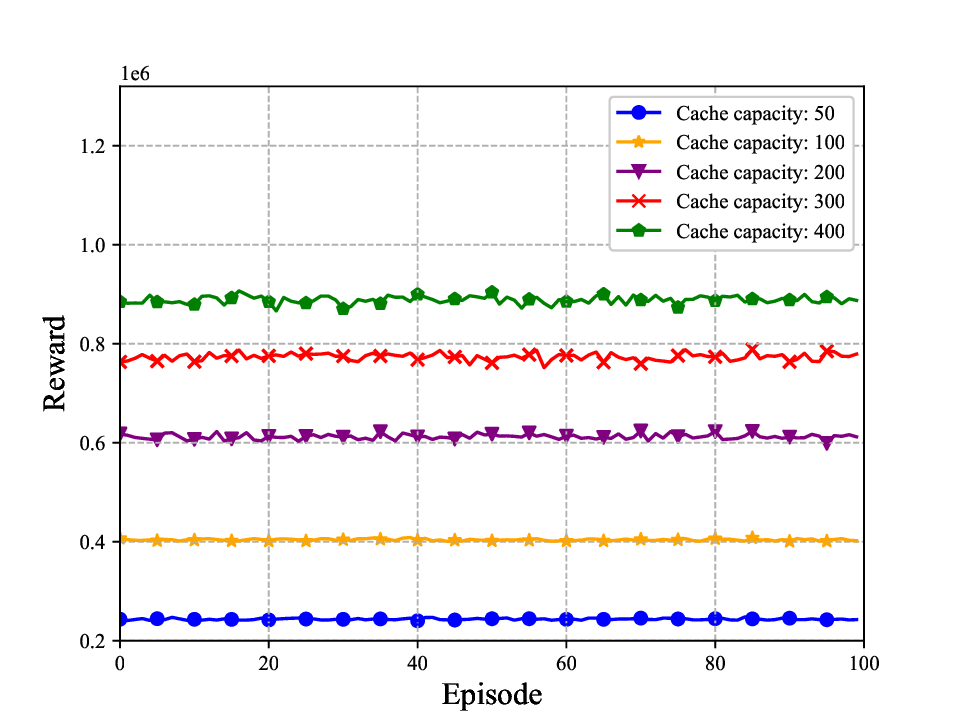}
    \end{minipage}}
  \subfigure[]{
    \begin{minipage}{7.5cm}
    \includegraphics[scale=0.45]{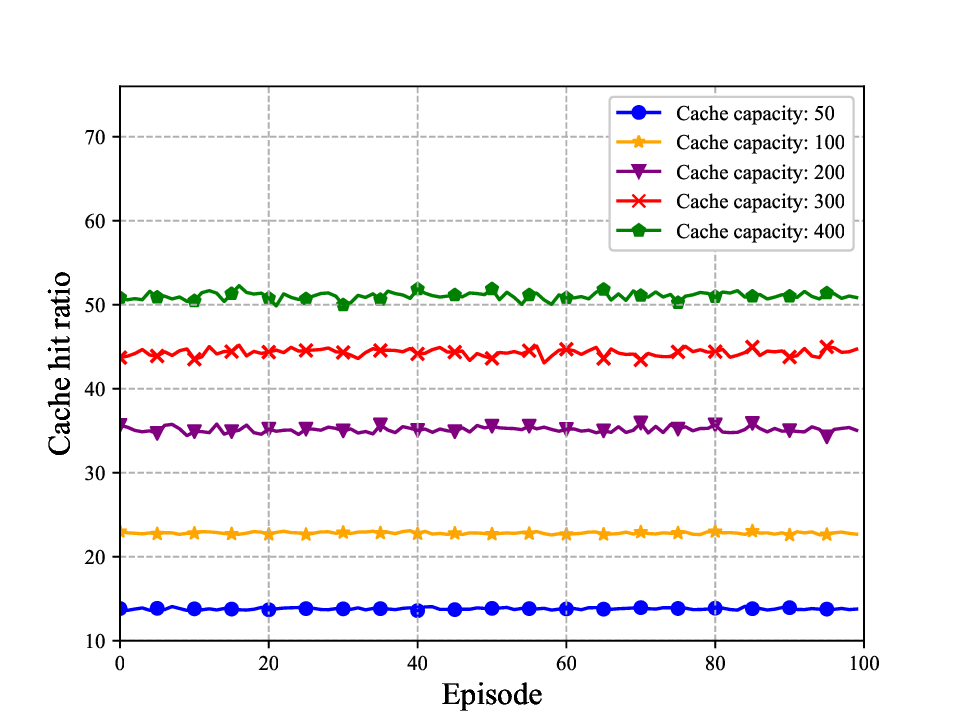}
    \end{minipage}}
    \caption{Testing performance vs episode for different cache capacities. (a) Cost; (b) Reward; (c) Cache hit ratio.}
    \label{fig8}
    \vspace{-0.5cm}
\end{figure}

\begin{figure}[h]
  \centering
  \subfigure[]{
    \begin{minipage}{7.5cm}
    \includegraphics[scale=0.45]{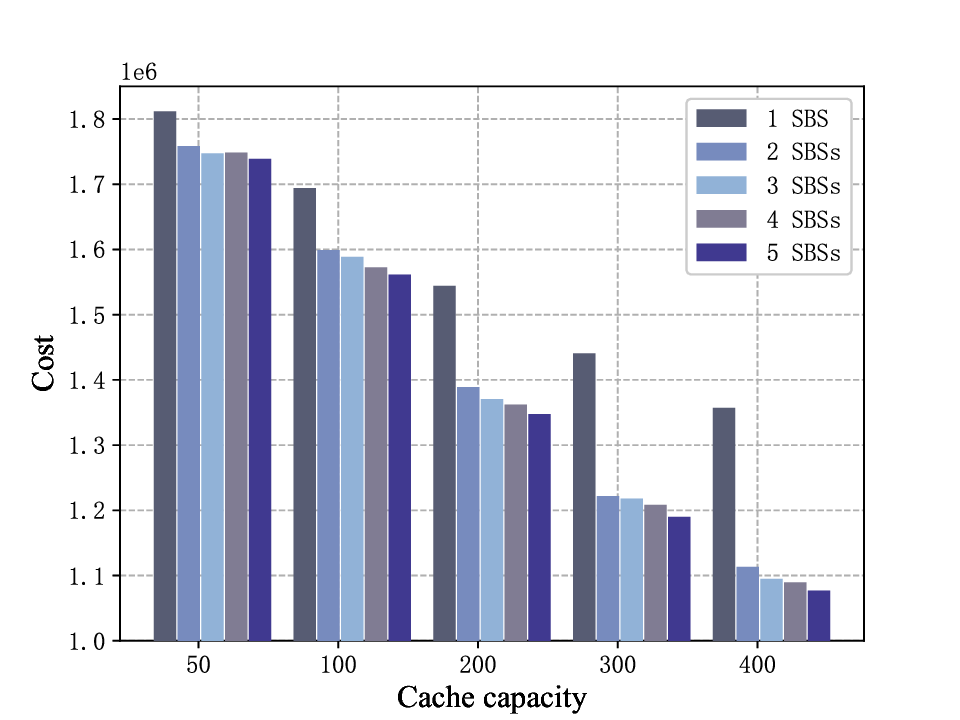}
    \end{minipage}}
  \subfigure[]{
    \begin{minipage}{7.5cm}
    \includegraphics[scale=0.45]{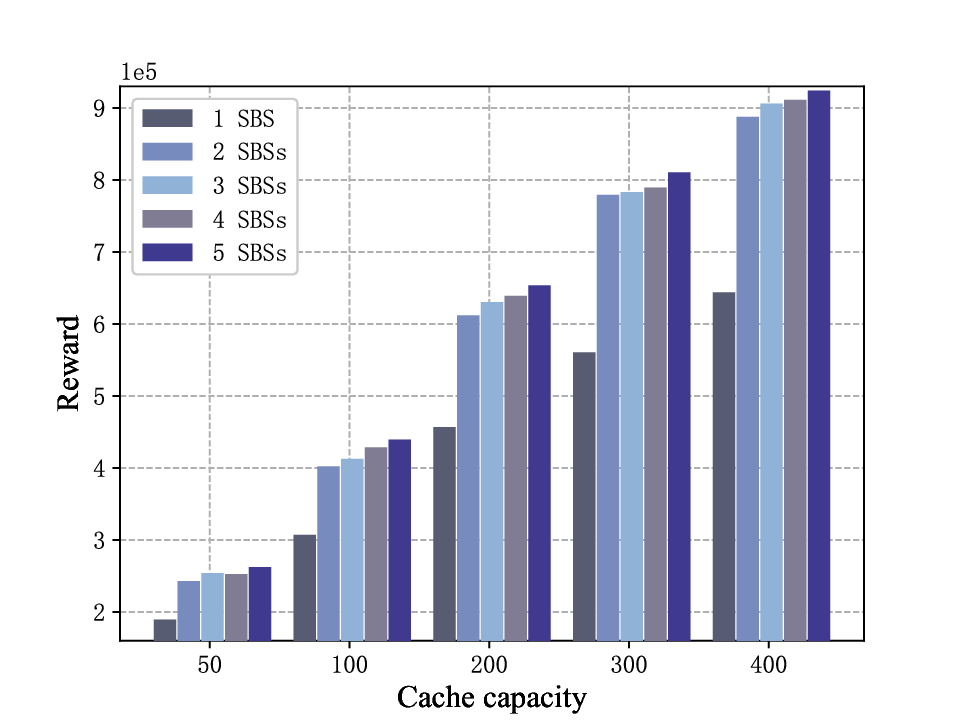}
    \end{minipage}}
  \subfigure[]{
    \begin{minipage}{7.5cm}
    \includegraphics[scale=0.45]{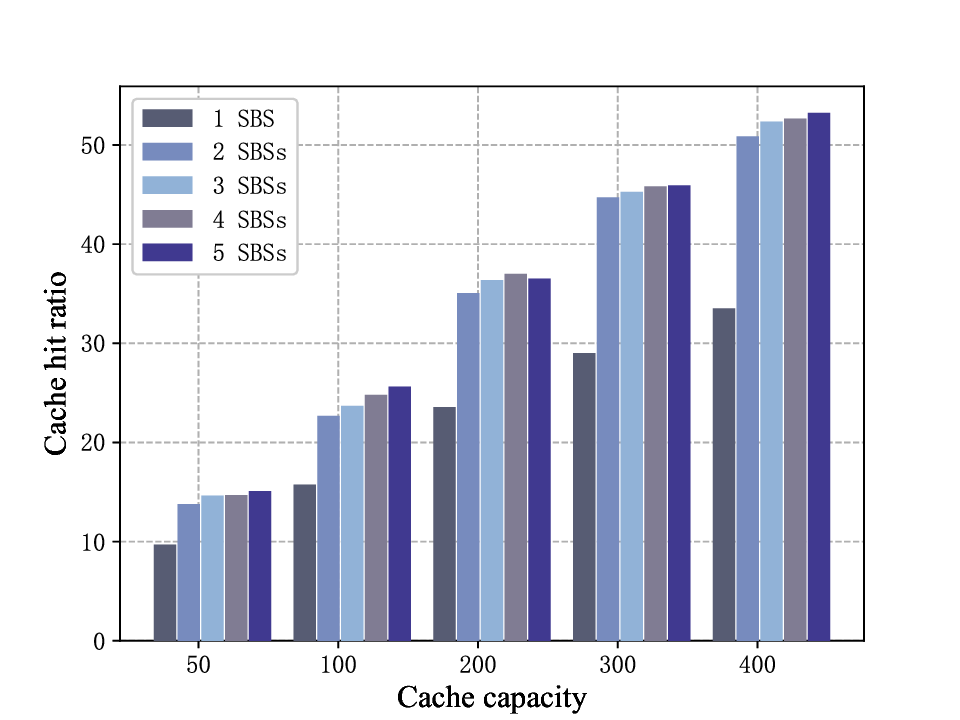}
    \end{minipage}}
    \caption{Performance vs cache capacity for different number of SBSs. (a) Cost; (b) Reward; (c) Cache hit ratio.}
    \label{fig6}
    \vspace{-0.5cm}
\end{figure}


\subsection{Performance Evaluation}
The experiments use the cost and cache hit ratio of the network as performance metrics. The cache hit ratio quantifies the utilization of the edge caching.
If the content requested by a UE within the coverage of a local SBS $b$ is cached in the SBS or adjacent SBSs, it is called a cache hit. Otherwise, it is called a cache miss.
	Thus the cache ratio of SBS $b$ can be calculated as
\begin{equation}
CH_b=\frac{\text { cache hits }}{\text { cache misses + cache hits }}.
\label{eq29}
\end{equation}
The average cache hit ratio of all SBSs is calculated as

\begin{equation}
C H=\frac{1}{B} \sum_{b=1}^B C H_b.
\label{eq28}
\end{equation}

In order to evaluate the performance of the proposed CEFMR scheme, three baseline caching schemes are introduced as below for comparison.

\begin{itemize}
\item Random: Each SBS randomly selects $C$ contents from all contents for caching.

\item C-$\varepsilon$-greedy:
Each SBS caches the $C$ contents with the highest number of requests based on the probability of $1-\varepsilon$, it randomly selects $C$ contents for caching with a probability of $\varepsilon$. In the simulation, $\varepsilon = 0.1$.

\item Thompson sampling: Each SBS updates the contents cached in that SBS based on the number of cache hits and cache misses in the previous time slot, and caches the $C$ contents with the highest values.
\item Bayes-Based Popularity Learning and Sequential Greedy Algorithm (BSG)\cite{9128365}: The algorithm first initializes each content’s popularity with a Beta distribution and updates their distribution based on the observed frequency of requests and non-requests for each content. Then the algorithm sorts SBS based on the sum of the popularity of UE requested contents within the SBS coverage, meanwhile it sorts all contents based on popularity. It then allocates most popular contents to the top-sorted SBS.
\item Traditional Federated MADRL (TFMADRL): This scheme employs traditional federated learning to train the AAE model, and then utilizes MADDPG for collaborative caching.
\item Elastic Federated learning with Non-RL (EFNRL): This scheme utilizes elastic federated learning to train the AAE model. During caching, each SBS sorts the predicted popularity of content and then caches the top $C$ contents.
\end{itemize}

For the Random scheme, the complexity is \(O(C)\), where \(C\) is the cache capacity of the SBS. The C-\(\varepsilon\)-greedy scheme's complexity is \(O(B \times m \log m)\), with \(B\) representing the number of UEs covered by each SBS, and \(m\) being the number of content requests by UEs. Thompson sampling has an approximate complexity of \(O(e_t \times B \times m \log m)\). The BSG scheme's complexity is approximately \(O(m_{all} \log m_{all})\), where \(m_{all}\) indicates all cached content in the CS. The complexity of both Traditional Federated MADRL and our CEFMR is approximated as \(O_{aarfl} + O_{madrl}\). The complexity of EFNRL is approximately \(O_{aarfl}\).

Figs. \ref{fig9}-(a), (b) and (c) show the average weight distance of each layer between UEs' local AAE models and global AAE model, and the average weight distance of all layers between UEs' local AAE models and global AAE model under different rounds in elastic FL training. It can be seen that all the weight distances tend to decrease as round increases. After a certain number of rounds, all the weight distances converge. This is because as the training of elastic FL, both UEs' local models and global model gradually converge, and the local models are trained based on the downloaded global model, the divergence between local models and global model will gradually decrease. It also can be seen that all weight distances do not reduce to zero after convergence. This is because the data among UEs are diverse, thus the models trained locally by UEs maintain their own characteristics.

Figs. \ref{fig5}-(a) and (b) show the total cost of all SBSs, where the cost of each SBS $b$ is $\alpha r_b^t+\beta r_{b, n}^t+\chi r_{b, c}^t+\delta r_{b, e}^t$, and cache hit ratio with different cache capacities of each SBS under different schemes, respectively. It can be seen that the costs of all schemes decrease as the cache capacity increases, and the cache hit ratios increase as the cache capacity increases.
This is because all SBSs which have larger caching capacities can cache more contents. As a result, UEs can more easily fetch their requested contents from SBSs. This can alleviate the pressure on the CS and improve the caching performance of the system.
The Random and Thompson Sampling schemes have poor performance. This is because these schemes do not cache contents based on predicting content popularity, while other five schemes determine the cached contents based on the predicted interested contents uploaded by the UEs within the coverage of SBS. Both CEFMR and TFMADRL schemes outperform EFNRL, BSG, and C-$\varepsilon$-greedy schemes. This is because CEFMR and TFMADRL schemes employ cooperative caching using the MADDPG method. CEFMR and TFMADRL schemes also outperform BSG scheme because they utilize a more efficient AAE model for popularity prediction. Furthermore, from the figure, it can be observed that as cache capacity increases, CEFMR scheme gradually outperforms TFMADRL scheme. This is because CEFMR scheme employs elastic federated learning to train the AAE, which allows individual UE's personalized models to better reflect their own data. This means that for each UE, the personalized AAE model is more likely to provide more accurate predictions of content popularity. When the SBS's cache capacity increases, more accurate predictions can help cache more suitable content, thereby increasing cache hit rates and reducing costs.

Figs. \ref{fig8}-(a), (b) and (c) show the fluctuation figures of the total cost of all SBSs, reward and cache hit ratio with respect to episodes in the testing stage of MADDPG under different cache capacities for each SBS. It can be seen that the performance of the MADDPG tends to be smooth in different testing episodes, which demonstrates the stability of the algorithm. From these figures, it can also be seen that cost decreases as cache capacity increases, while reward and cache hit ratio increase as cache capacity increases. This is because as the cache capacity increases, SBS can cache more contents requested by UEs, and UEs can fetch more contents from the local SBS, thereby improving the cache hit rate. Moreover, as cache capacity increases, UEs can fetch more contents from the local SBSs or from adjacent SBSs, instead of fetching them from CS. Therefore, increasing cache capacity can reduce cost and increase the saved cost, i.e., the reward.

Figs \ref{fig6}-(a), (b), and (c) show the variations in cost, reward, and cache hit ratio with respect to the cache capacity of each SBS under different numbers of SBSs.
When the number of SBSs is $1$, this case provides the worst performance. This is because the CEFMR scheme has become a single-agent scheme. The local SBS cannot obtain requested contents from the adjacent SBSs, thus the UEs can only fetch the requested contents from the local SBS or CS with a higher cost.
As the number of SBSs increases, the CEFMR scheme generally performs better in terms of cost, reward, and cache hit ratio. This is because more SBSs can cooperate to cache popular contents. It also indicates the effectiveness of the CEFMR scheme in coordinating edge caching.

\section{Conclusions}
\label{sec7}
In this paper, we proposed a cooperative edge caching scheme named CEFMR to optimize the cost to fetch contents in next-generation networks. We first proposed an elastic FL algorithm to train the personalized model for each UE, where the AAE model was adopted for training to improve the prediction accuracy, then a popular content prediction algorithm was proposed to predict the popular contents for each SBS based on the trained AAE model. Finally, we proposed a MADRL based algorithm to decide where the predicted popular contents are collaboratively cached among SBSs to reduce the cost for fetching contents. Our experimental results demonstrated the superiority of our proposed scheme to existing baseline caching schemes. The conclusions are summarized as follows:
\begin{itemize}

	\item{CEFMR's ability to extract hidden features from the local data of UEs significantly outperforms baseline schemes in predicting popular content for caching. Unlike other schemes, CEFMR's personalized approach personalizes predictions to the specific usage patterns and preferences of each UE. This leads to a more accurate and efficient caching process, as evidenced by a comparison without AAE model, which shows the improvements in the cache hit ratio.}
	
	\item{The elastic FL algorithm within CEFMR uniquely assigns specific weights to the model of each UE based on the distance between previously trained local models and the global model. This personalization ensures that each UE's model accurately reflects its data characteristics, a feature not commonly seen in other schemes like traditional FL. This approach has demonstrated CEFMR can enhance the efficiency of the cached content and save cost.}
	
	\item{The MADDPG algorithm in CEFMR facilitates cooperative decision-making on whether to cache content in local SBS, adjacent SBSs or CS. This not only optimizes cache storage distribution but also significantly reduces operational costs compared to the schemes without MADRL. Our comparative analysis reveals that this method leads to improvements in cache efficiency and cost savings.}
	
\end{itemize}

\ifCLASSOPTIONcaptionsoff
  \newpage
\fi

\bibliographystyle{IEEEtran}

%
%

\vspace{-1cm}
\begin{IEEEbiography}[{\includegraphics[width=1in,height=1.25in,clip,keepaspectratio]{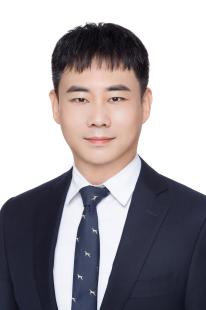}}] {Qiong Wu} (Senior Member, IEEE) received the Ph.D. degree in information and communication engineering from National Mobile Communications Research Laboratory, Southeast University, Nanjing, China, in 2016. From 2018 to 2020, he was a postdoctoral researcher with the Department of Electronic Engineering, Tsinghua University, Beijing, China. He is currently an associate professor with the School of Internet of Things Engineering, Jiangnan University, Wuxi, China. Dr. Wu is a senior member of IEEE. 
He has published over 60 papers in high impact journals and conferences and authorized over 20 patents. He has severed as section board member of Sensors, early career editorial board member of CMC-Computers Materials \& Continua and Chinese Journal on Internet of Things, lead guest editor of Sensors, CMC-Computers Materials \& Continua and Frontiers in Space Technologies, guest editor of Electronics, TPC co-chair of WCSP'22, TPC member of EIIE'24, ICNC'24, ICCT'23, VTC'22 fall, VTC'20 fall, ICNC'20 and session chair of WCSP'22, ICICN'21, WCSP'20. 
His current research interest focuses on vehicular networks, autonomous driving communication technology, and machine learning.
\end{IEEEbiography}

\begin{IEEEbiography}[{\includegraphics[width=1in,height=1.25in,clip,keepaspectratio]{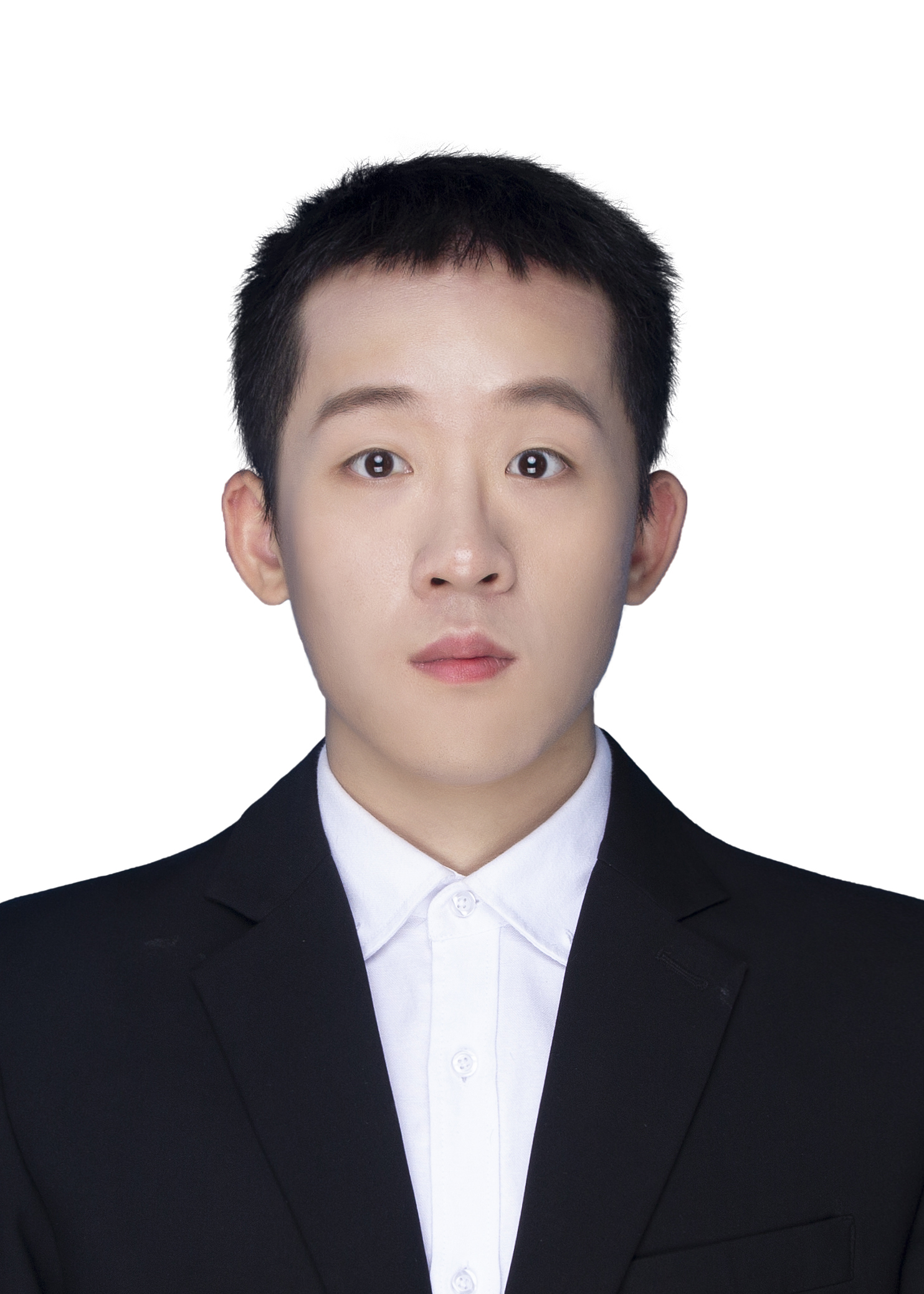}}] {Wenhua Wang} received the B.S. degree from the Jiangnan University, Wuxi, China, in 2021. He is currently working toward the M.S. degree with Jiangnan University. His research interests include federated learning, reinforcement learning, and mobile edge computing.
\end{IEEEbiography}

\begin{IEEEbiography}[{\includegraphics[width=1in,height=1.25in,clip,keepaspectratio]{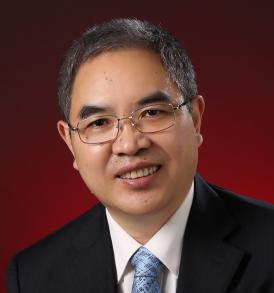}}] {Pingyi Fan} (Senior Member, IEEE) received the B.S. degree from the Department of Mathematics, Hebei University, in 1985, the M.S. degree from the Department of Mathematics, Nankai University, in 1990, and the Ph.D. degree from the Department of Electronic Engineering, Tsinghua University, Beijing, China, in 1994. 
From August 1997 to March 1998, he visited The Hong Kong University of Science and Technology as a Research Associate. From May 1998 to October 1999, he visited the University of Delaware, Newark, DE, USA, as a Research Fellow. In March. 2005, he visited NICT, Japan, as a Visiting Professor. From June 2005 to May 2014, he visited The Hong Kong University of Science and Technology for many times. From July 2011 to September 2011, he was a Visiting Professor at the Institute of Network Coding, The Chinese University of Hong Kong.
He is currently a Professor and the director of open source data recognition innovation center at the Department of Electrical Engineering (EE), Tsinghua University. His main research interests include B5G technology in wireless communications, such as MIMO, OFDMA, network coding, network information theory, machine learning, and big data analysis. Dr. Fan is a Fellow of IET and an Overseas Member of IEICE. He is also a reviewer of more than 40 international journals, including 30 IEEE journals and eight EURASIP journals. 
He has received some academic awards, including the IEEE WCNC'08 Best Paper Award, the IEEE ComSoc Excellent Editor Award for IEEE TRANSACTIONS ON WIRELESS COMMUNICATIONS in 2009, the ACM IWCMC'10 Best Paper Award, the IEEE Globecom'14 Best Paper Award, the IEEE ICC'20 Best Paper Award, the IEEE TAOS Technical Committee'20 Best Paper Award,  IEEE ICCCS Best Paper Awards in 2023 and 2024, the CIEIT Best Paper Awards in 2018 and 2019. He has served as an editor of IEEE Transactions on Wireless Communications, Inderscience International Journal of Ad Hoc and Ubiquitous Computing and Wiley Journal of Wireless Communication and Mobile Computing, MDPI Electronics and Open Journal of Mathematical Sciences etc. 
He has attended to organize many international conferences, as the general chairs/TPC chair/ Plenary/Keynote speaker for over 30 international Conferences, including as the General Co-Chair of EAI Chinacom2020, 2023 and IEEE VTS HMWC 2014, the TPC Co-Chair of IEEE ICCCS2024 and a TPC Member of IEEE ICC, Globecom, WCNC, VTC, and Inforcom. He has served as an Editor for IEEE TRANSACTIONS ON WIRELESS COMMUNICATIONS, International Journal of Ad Hoc and Ubiquitous Computing (Inderscience), Journal of Wireless Communication and Mobile Computing (Wiley), Electronics (MDPI), and Open Journal of Mathematical Sciences, IAES international journal of artificial intelligence. He is also an associate editor of IEEE Transactions on cognitive communications and networking.
\end{IEEEbiography}


\begin{IEEEbiography}[{\includegraphics[width=1in,height=1.25in,clip,keepaspectratio]{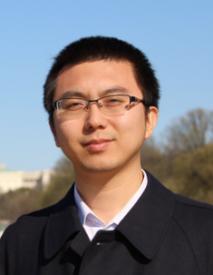}}] {Qiang Fan} received his Ph.D. degree in Electrical and Computer Engineering from New Jersey Institute of Technology (NJIT) in 2019, and his M.S. degree in Electrical Engineering from Yunnan University of Nationalities, China, in 2013. He was a postdoctor researcher in the Department of Electrical and Computer Engineering, Virginia Tech. Currently, he is a staff engineer in Qualcomm, USA. He has served as a reviewer for over 120 journal submissions such as IEEE Transactions on Cloud Computing, IEEE Journal on Selected Areas in Communications, IEEE Transactions on Communications. His current research interests include wireless communications and networking, mobile edge computing, machine learning and drone assisted networking.\end{IEEEbiography}

\begin{IEEEbiography}[{\includegraphics[width=1in,height=1.25in,clip,keepaspectratio]{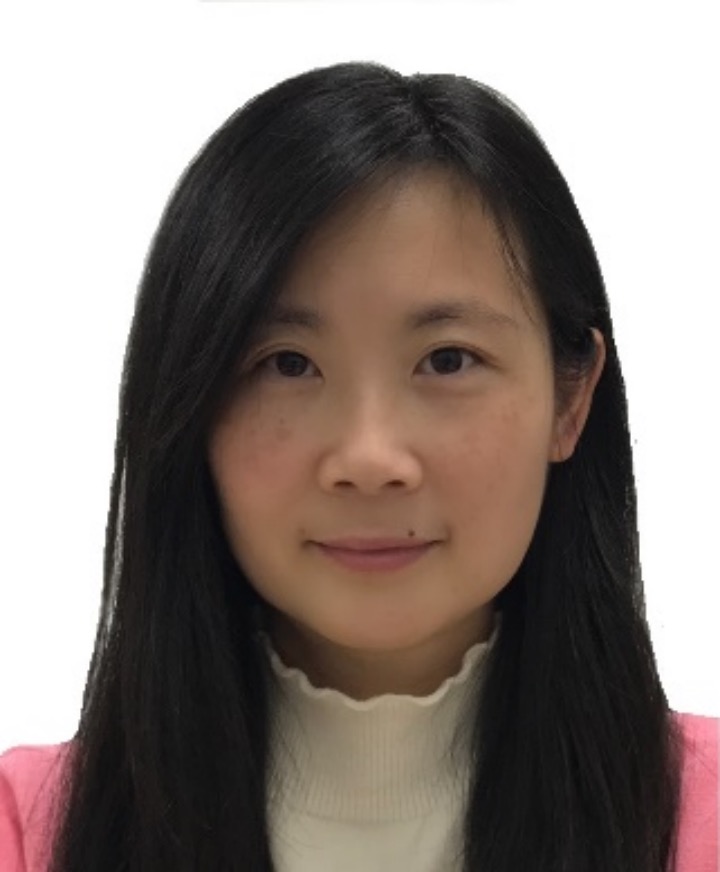}}] {JHuling Zhu} (SM'17) received the B.S degree from Xidian University, China, and the Ph.D. degree from Tsinghua University, China. She is currently a Reader (Associate Professor) in the School of Engineering, University of Kent, United Kingdom. Her research interests are in the area of wireless communications. She was holding European Commission Marie Curie Fellowship from 2014 to 2016. She received the best paper award from IEEE Globecom 2011. She was Symposium Co-Chair for IEEE Globecom 2015 and IEEE ICC 2018, and Track Co-Chair of IEEE VTC2016-Spring and VTC2018-Spring. Currently, she serves as an Editor for IEEE Transactions on Vehicular Technology.\end{IEEEbiography}

\begin{IEEEbiography}[{\includegraphics[width=1in,height=1.25in,clip,keepaspectratio]{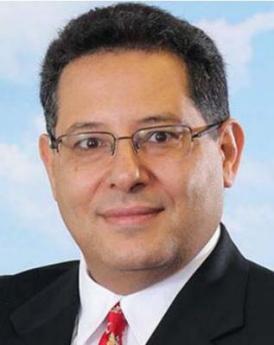}}] {Khaled Ben Letaief} (Fellow, IEEE) received the B.S. (Hons.), M.S., and Ph.D. degrees in electrical engineering from Purdue University, West Lafayette, IN, USA, in December 1984, August 1986, and May 1990, respectively. 
From 1990 to 1993, he was a Faculty Member with The University of Melbourne, Melbourne, VIC, Australia. 
Since 1993, he has been with The Hong Kong University of Science and Technology (HKUST). 
While at HKUST, he has held many administrative positions, including the Head of the Electronic and Computer Engineering Department, the Director of the Wireless IC Design Center, the Founding Director of Huawei Innovation Laboratory, and the Director of Hong Kong Telecom Institute of Information Technology. While at HKUST, he has also served as a Chair Professor and the Dean of Engineering. Under his leadership, the School of Engineering has not only transformed its education and scope and produced very high caliber scholarship, but it has also actively pursued knowledge transfer and societal engagement in broad contexts. 
It has also dazzled in international rankings (rising from \#26 in 2009 to \#14 in the world in 2015 according to QS World University Rankings). From September 2015 to March 2018, he joined HBKU as a Provost to help establish a research-intensive university in Qatar in partnership with strategic partners that include Northwestern University, Carnegie Mellon University, Cornell, and Texas A\&M. He served as Consultants for different organizations, including Huawei, ASTRI, ZTE, Nortel, PricewaterhouseCoopers, and Motorola. He is currently with the Peng Cheng Laboratory, Shenzhen, China. 
He is also an Internationally Recognized Leader in wireless communications and networks with research interests in artificial intelligence, big data analytics systems, mobile cloud, and edge computing, tactile internet, and 5G systems. 
In these areas, he has over 630 articles with over 38,350 citations and an H-index of 87 along with 15 patents, including 11 U.S. inventions. 
Dr. Letaief is a member of the United States National Academy of Engineering, a fellow of Hong Kong Institution of Engineers, and a member of Hong Kong Academy of Engineering Sciences. 
He was a recipient of many distinguished awards and honors, including the 2019 Distinguished Research Excellence Award by HKUST School of Engineering (Highest Research Award and only one recipient/three years is honored for his/her contributions), the 2019 IEEE Communications Society and Information Theory Society Joint Paper Award, the 2018 IEEE Signal Processing Society Young Author Best Paper Award, the 2017 IEEE Cognitive Networks Technical Committee Publication Award, the 2016 IEEE Signal Processing Society Young Author Best Paper Award, the 2016 IEEE Marconi Prize Paper Award in Wireless Communications, the 2011 IEEE Wireless Communications Technical Committee Recognition Award, the 2011 IEEE Communications Society Harold Sobol Award, the 2010 Purdue University Outstanding Electrical and Computer Engineer Award, the 2009 IEEE Marconi Prize Award in Wireless Communications, the 2007 IEEE Communications Society Joseph LoCicero Publications Exemplary Award, and more than 16 IEEE best paper awards. 
He is well recognized for his dedicated service to professional societies and IEEE, where he has served in many leadership positions, including the Treasurer of the IEEE Communications Society, the IEEE Communications Society VicePresident for Conferences, the Chair of the IEEE Committee on Wireless Communications, an Elected Member of the IEEE Product Services and Publications Board, and the IEEE Communications Society Vice-President for Technical Activities. 
He is the Founding Editor-in-Chief of the prestigious IEEE TRANSACTIONS ON WIRELESS COMMUNICATIONS and has served on the Editorial Board of other premier journals, including the IEEE JOURNAL ON SELECTED AREAS IN COMMUNICATIONS: Wireless Communications Series (as the Editor-in-Chief). 
He has also been involved in organizing many flagship international conferences. He also served as the President of the IEEE Communications Society from 2018 to 2019, the world's leading organization for communications professionals with headquarter in New York City and members in 162 countries. 
He is also recognized by Thomson Reuters as an ISI Highly Cited Researcher and was listed among the 2020 top 30 of AI 2000 Internet of Things Most Influential Scholars. IEEE Committee on Wireless Communications, an Elected Member of the IEEE Product Services and Publications Board, and the IEEE Communications Society Vice-President for Technical Activities. 
He is the Founding Editor-in-Chief of the prestigious IEEE TRANSACTIONS ON WIRELESS COMMUNICATIONS and has served on the Editorial Board of other premier journals, including the IEEE JOURNAL ON SELECTED AREAS IN COMMUNICATIONS: Wireless Communications Series (as the Editor-in-Chief). 
He has also been involved in organizing many flagship international conferences. He also served as the President of the IEEE Communications Society from 2018 to 2019, the world's leading organization for communications professionals with headquarter in New York City and members in 162 countries. 
He is also recognized by Thomson Reuters as an ISI Highly Cited Researcher and was listed among the 2020 top 30 of AI 2000 Internet of Things Most Influential Scholars.

\end{IEEEbiography}
\end{document}




\ifCLASSOPTIONcaptionsoff
  \newpage
\fi

\end{document}